\documentclass{article}

 \usepackage[main, final]{neurips_2026}

\usepackage[utf8]{inputenc} 
\usepackage[T1]{fontenc}    
\usepackage{hyperref}       
\usepackage{url}            
\usepackage{booktabs}       
\usepackage{amsfonts}       
\usepackage{nicefrac}       
\usepackage{microtype}      
\usepackage{xcolor}         
\usepackage{algorithm}
\usepackage{algpseudocode}
\usepackage{amsmath, amssymb}

\usepackage{multirow} 
\usepackage{amssymb}  

\usepackage{xcolor}  
\usepackage{colortbl} 
\usepackage{graphicx}
\usepackage{subcaption}
\usepackage{wrapfig}
\usepackage{booktabs} 
\usepackage{multirow}  
\usepackage{makecell}
\usepackage{graphicx} 
\usepackage{amsmath} 
\usepackage{enumitem}
\usepackage{tabularx}
\usepackage{paracol}
\usepackage{listings}
\usepackage{textcomp} 

\usepackage[most]{tcolorbox} 
\tcbuselibrary{raster} 
\usepackage{xcolor}
\usepackage{soul} 
\usepackage{enumitem}
\usepackage{pifont}
\usepackage{helvet}
\definecolor{unfaithfulbg}{RGB}{240,248,250}

\definecolor{obsPurple}{RGB}{128, 0, 128}      
\definecolor{planblue}{RGB}{0, 0, 200}         
\definecolor{failred}{RGB}{220,80,80}   
\definecolor{passgreen}{RGB}{90, 160, 90} 
\definecolor{SoftGray}{HTML}{7F7F7F}      
\definecolor{SoftRed}{HTML}{D9534F}       
\definecolor{SoftGreen}{HTML}{5CB85C}     

\definecolor{BrightRed}{HTML}{FF6B6B} 
\definecolor{BrightGreen}{HTML}{4CAF50}     

\algnewcommand{\algorithmicinput}{\textbf{Input:}}
\algnewcommand{\algorithmicoutput}{\textbf{Output:}}
\algnewcommand{\INPUT}{\item[\algorithmicinput]}
\algnewcommand{\OUTPUT}{\item[\algorithmicoutput]}

\usepackage{caption}
\newcommand{\cmark}{\textcolor{green!60!black}{\checkmark}} 
\newcommand{\xmark}{\textcolor{red}{\times}}               

\title{\Large MCPEvol-Bench: Benchmarking LLM Agent Performance Across Dynamic Evolutions of MCP Servers}

\author{
    Huanxi Liu$^{1,2,3}$ \quad
    Kun Hu$^{1}$ \quad
    Jiaqi Liao$^{1,2,3}$ \quad
    Qiang Wang$^{1,2,3}$ \quad
    Pengfei Qian$^{1,2,3}$ \\[0.5ex]
    \textbf{YuanZhao Zhai}$^{1,2,3}$ \quad
    \textbf{Dawei Feng}$^{1,2,3}$\thanks{Correspondence to: \texttt{dafeng@nudt.edu.cn}} \quad
    \textbf{Bo Ding}$^{1,2,3}$ \quad
    \textbf{Huaimin Wang}$^{1,2,3}$ \\[1.5ex]
    $^{1}$ College of Computer Science and Technology, National University of Defense Technology \\
    $^{2}$ State Key Laboratory of Complex \& Critical Software Environment \\ 
    $^{3}$ National Key Laboratory of Parallel and Distributed Computing
}

\begin{document}
\maketitle

\begin{abstract}
As Model Context Protocol (MCP) servers emerge as the core infrastructure for connecting LLMs with external tools, existing benchmarks leverage real-world MCP servers to evaluate LLM agents' tool-using capabilities.
However, these benchmarks overlook the continuous evolution of tool interfaces and functionalities within MCP servers, resulting in flawed assessments that fail to capture the agent's adaptability in changing tool landscapes.
To bridge this gap, we introduce \textbf{MCPEvol-Bench}, a novel benchmark for evaluating the task-solving capabilities of LLM agents under dynamic toolset evolution.
Inspired by large-scale empirical study, we propose 11 mutation operators to simulate realistic tool evolution within 123 MCP servers.
We benchmark 12 state-of-the-art LLMs on multiple versions of MCP servers, revealing that even frontier models struggle to adapt to evolving tools.
For instance, GPT-5.4 and Claude-Sonnet-4-6 exhibit performance declines of 13.7\% and 14.4\% in evolved MCP servers, respectively, accompanied by substantial increases in planning and reasoning errors.
These findings highlight the vulnerability of LLM-driven workflows, establishing MCPEvol-Bench as a standard for evaluating agent adaptability in dynamic tool environments.
\end{abstract}

\section{Introduction}
Recent advances in large language models (LLMs) have inspired widespread efforts to develop tool-using agents capable of comprehending natural language instructions, planning multi-step workflows, and interacting with external tools to solve complex tasks~\cite{agentsurvey, toolsurvey, agenticsurvey}. 
These agents have been widely adopted in various real-world domains, such as software development~\cite{repomaster}, system automation~\cite{OS-ATLAS}, and scientific research~\cite{ChemAgent}, where LLM agents address user tasks by chaining multiple tools and orchestrating interdependent operations.

Despite significant progress, existing benchmarks for evaluating LLM agents’ tool-using capabilities remain fundamentally limited. 
Early efforts~\cite{toolllm, toolalpaca, shortcutsbench} provided unstable evaluations by assessing agents through interactions with real-world APIs, which suffered from frequent endpoint changes and service deprecations~\cite{stabletoolbench}.
Recently, the Model Context Protocol (MCP)~\cite{anthropic2024mcp} has emerged as a unified standard for connecting LLMs with external tools, leveraging dynamic server discovery to overcome the inflexibility of prior API-based integration.
However, as summarized in Table \ref{tab:comparison}, current MCP-based benchmarks overlook the continuous evolution of tool interfaces and functionalities within MCP servers.
Consequently, these benchmarks cannot accurately evaluate the adaptability of LLM agents in preserving the integrity of original workflow under real-world tool changes. 
\begin{figure*}[t]
    \vspace{-0.5em}
    \centering
    \includegraphics[width=0.85\linewidth]{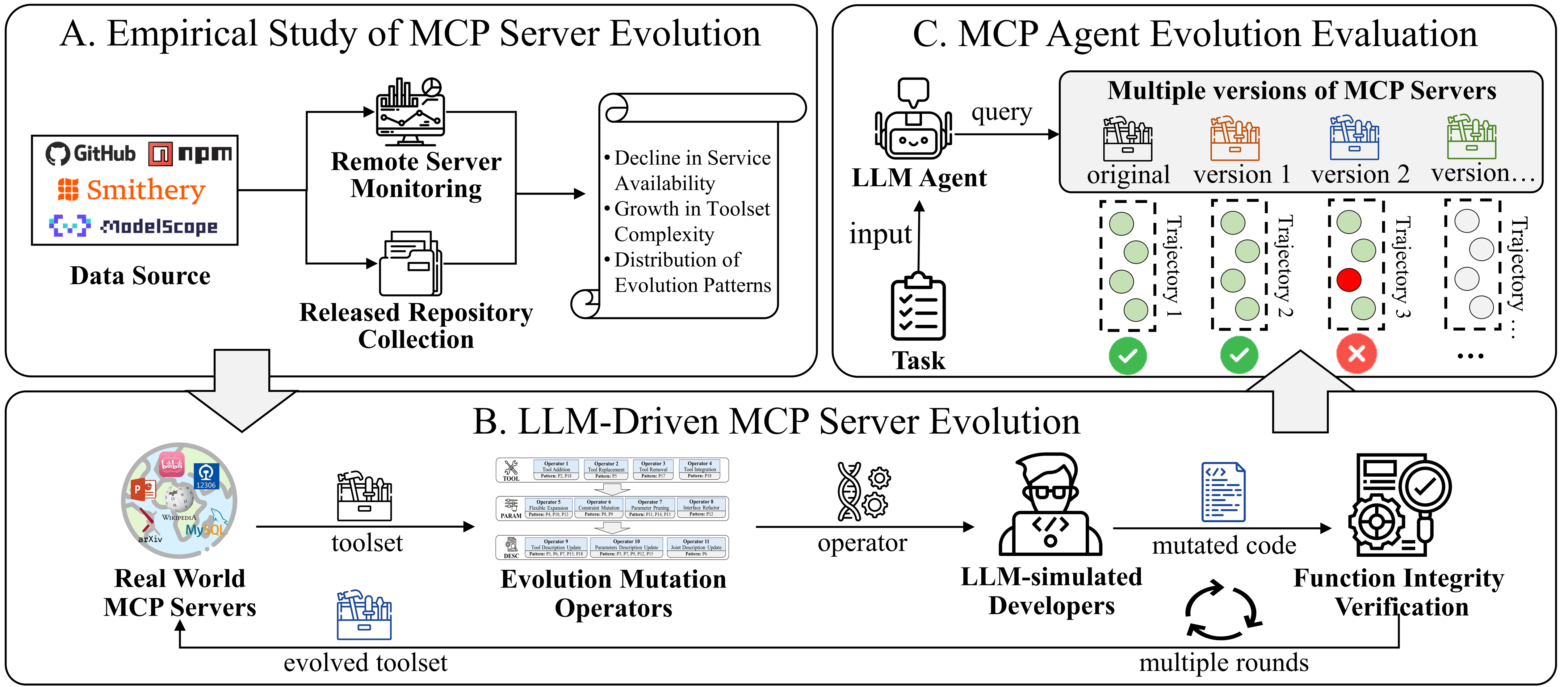} 
    \caption{
         The framework of MCPEvol-Bench. 
         \textbf{(A) Empirical Study of MCP Server Evolution}: Validates the prevalence of evolution and identifies predominant evolution patterns. 
         \textbf{(B) LLM-Driven MCP Server Evolution}: Simulates developer behavior to adaptively select mutation operators for source code modification, generating evolved toolsets. 
         \textbf{(C) MCP Agent Evolution Evaluation}: Assesses the task performance of LLM agents across multiple versions of MCP servers.
    }
    \label{fig:overview}
\end{figure*}

\begin{table}[b]
\vspace{-0.3cm}
\centering
\caption{Comparison of representative tool-using benchmarks.}
\label{tab:comparison}
\resizebox{\columnwidth}{!}{
\begin{tabular}{l c c c  c c c} 
\toprule
\textbf{Benchmark} & \small\textbf{\# Domain} & \small\textbf{\# Tool/API} & \small\textbf{\makecell{MCP Ecosystem}} & \small\textbf{\makecell{Real-World \\ Toolset}} & \small\textbf{\makecell{Multi-Tool \\ Collaboration}} & \small\textbf{\makecell{Dynamic \\ Evolution}} \\ 
\midrule
ToolBench~\cite{toolllm}       & 19 & 3451 & $\xmark$ & $\cmark$ & $\cmark$ & $\xmark$ \\
SHORTCUTSBENCH~\cite{shortcutsbench}      & 50 & 426  & $\xmark$ & $\cmark$ & $\cmark$ & $\xmark$  \\
APIGen~\cite{APIGEN}           & 49 & 3673 & $\xmark$ & $\xmark$ & $\cmark$ & $\xmark$ \\
MCP-Flow~\cite{mcpflow} & 10 & 11536 & $\cmark$ & $\cmark$ & $\xmark$ & $\xmark$ \\
MCPEval~\cite{mcpeval}             & 5 & 19 & $\cmark$ & $\cmark$ & $\cmark$ & $\xmark$ \\
MCP-Bench~\cite{mcp_bench}   & 28 & 250 & $\cmark$ & $\cmark$& $\cmark$ & $\xmark$ \\
MCPEvol-Bench (Ours)               & 9 & 1272 & $\cmark$ & $\cmark$ & $\cmark$ & $\cmark$ \\ 
\bottomrule
\end{tabular}
}
\vspace{-0.5em}
\end{table}

To bridge this gap, we introduce \textbf{MCPEvol-Bench}, a novel benchmark for evaluating the task-solving capabilities of LLM agents under dynamic toolset evolution, as illustrated in Figure \ref{fig:overview}. 
We first conduct a comprehensive empirical study of MCP servers by analyzing temporal changes in remotely hosted servers and version histories of public code repositories.
This study reveals the prevalence and patterns of MCP server evolution, with 20.7\% of remote servers unavailable and 54.6\% of tools in repository deleted or replaced.
From the collected repositories, we curate 123 MCP servers with 1,272 tools across diverse domains (e.g., software development, data analytics), and automatically generate cross-server tasks via an LLM-based pipeline.
While historical versions of MCP servers provide rich evolution data, frequent failures due to broken dependencies make them unsuitable for benchmarking.
To address this, we extract 11 mutation operators covering the \texttt{Tool}, \texttt{Parameter}, and \texttt{Description} levels from observed evolution patterns. 
Leveraging these operators as prompt instructions, we implement \textbf{LLM-Driven MCP Server Evolution}, where LLMs automatically select appropriate modifications and iteratively modify tools, thereby generating a series of evolved MCP servers.
Finally, in \textbf{MCP Agent Evolution Evaluation}, LLM agents execute tasks across multiple versions of MCP servers, measuring their task performance under dynamic toolset evolution.

In this research, we evaluate 12 representative LLMs on the benchmark, revealing that most widely used frontier models exhibit significant performance degradation on the evolved MCP servers. Notably, GPT-5.4 and Claude-Sonnet-4-6 suffer 13.7\% and 14.4\% drops in task fulfillment, respectively.
Analysis of agent trajectories shows that evolution primarily increases planning errors by 34.1\% and reasoning errors by 35.6\%.
By analyzing mutation operators, we find that the evolution involving tool additions or modifications severely impair agent performance, whereas removing redundant tools or parameters has  negligible effect on original workflow execution.
The reliability of our benchmark is validated by the high semantic similarity between simulated evolutions and real-world version updates, along with the strong agreement with human experts in evaluations.
Further experiments demonstrate that incorporating reflection, planning and memory modules enhances the adaptability of LLM agent systems, which highlights the critical role of cognitive components in evolving scenarios.

Our contributions can be summarized as follows:
\vspace{-0.1cm}
\begin{itemize}[leftmargin=2em, itemsep=1pt, topsep=0pt, parsep=0pt]
    \item We conduct a large-scale empirical study that validates the prevalence of MCP server evolution and uncovers its dominant patterns.
    Inspired by these patterns, we propose 11 mutation operators that empower LLMs to autonomously and iteratively perform precise MCP server evolution.
    \item We introduce MCPEvol-Bench, a novel benchmark comprising 123 multi-version MCP servers, which evaluates the task-solving capabilities of LLM agents under dynamic toolset evolution.
    \item Our extensive evaluation of 12 state-of-the-art LLMs across 201 challenging tasks reveals persistent weaknesses in dynamically evolving tool-using scenarios, highlighting the necessity of cognitive components to enhance LLM agent adaptability.
\end{itemize}

\begin{table*}[t]
    \centering
    \caption{An example from MCPEvol-BENCH, comparing original and evolved server configurations.}
    \label{tab:mcp_evolution_example}
    \footnotesize
\begin{tabularx}{\textwidth}{@{} 
    p{0.18\textwidth} 
    p{0.32\textwidth} 
    >{\raggedright\arraybackslash}X
@{}}
        \toprule
       \small{\textbf{Task}}  & \small{\textbf{Original Servers \& Tools}} & \small{\textbf{Evolved Servers \& Tools}}\\
        \midrule
        I need to plan a business trip from Beijing to Shanghai for next Wednesday.  \newline
        Please check the availability of high-speed train tickets in the morning, and then add them to my travel planning list in \texttt{work.md}. & 
        \textbf{Servers:} \textit{mcp-tasks, 12306-mcp}  \newline
        \textbf{Tools:} 
        \begin{itemize}[noitemsep, topsep=2pt, leftmargin=*]
            \item \texttt{tasks\_setup} ... 
            \item \texttt{tasks\_add} ... 
            \item \texttt{get-current-date} ... 
            \item \texttt{get-station-code-of-citys} ... 
            \item \texttt{get-tickets}: \{``fromStation'': the name or the station\_code of the departure location; ...\}
            \item \small{\texttt{......}}
        \end{itemize}
        & 
        \textbf{Servers:} \textit{mcp-tasks, 12306-mcp}  \newline
        \textbf{Tools:} 
        \begin{itemize}[noitemsep, topsep=2pt, leftmargin=*]
            \item \texttt{tasks\_setup} \textcolor{gray}{(unchanged)}; 
            \item \texttt{tasks\_add} \textcolor{gray}{(unchanged)}; 
            \item \texttt{get-current-date} \textcolor{gray}{(unchanged)};
            \item \texttt{get-station-code-of-citys} \textcolor{gray}{(unchanged)};
            \item \texttt{get-tickets} \textcolor{red}{(changed)}: \{``fromStation'': the name or the station\_code, \textcolor{red}{which can be queried by the get-station-code-of-citys}... , \textcolor{BrightGreen}{``minPrice''(new)}: ... \};
            \item \small{\texttt{......}}
        \end{itemize}
        \\
        \bottomrule
    \end{tabularx}
    \vspace{-0.3em}
\end{table*}

\section{Related Work}
\subsection{Tool-Use Benchmarks}
Integrating LLMs with external tools has emerged as a critical research direction for extending model capabilities beyond parametric knowledge. 
Early tool-use benchmarks, such as APIGEN~\cite{APIGEN} and ToolACE~\cite{toolace}, address API data scarcity through automated synthesis, while ToolBench~\cite{toolllm} and SHORTCUTSBENCH~\cite{shortcutsbench} aggregate numerous APIs from the web to enhance evaluation realism. 
However, real-world APIs are prone to frequent endpoint modifications and service deprecations, compromising evaluation stability~\cite{stabletoolbench}.
The emergence of the MCP provides a standardized invocation schema and stable service interfaces for building toolsets, facilitating the development of recent benchmarks like MCP-Bench~\cite{mcp_bench}, MCPEval~\cite{mcpeval}, and MCP-Flow~\cite{mcpflow}.
Despite this progress, existing MCP-based benchmarks assume a static tool environment, overlooking the dynamic nature of MCP servers, which undergo continuous evolution through version iterations. 
This limitation necessitates tool-use benchmarks that evaluate model performance under evolving server states.

\subsection{Evaluating LLM Agent Capability}
LLM agents are autonomous systems that leverage LLMs as their cognition core for decision-making, engaging in multi-turn tool interactions with external environments to accomplish complex user tasks~\cite{agentsurvey}. 
Existing evaluations typically target distinct agent capabilities: MCP-Bench~\cite{mcp_bench} assesses the handling of ambiguous instructions, ToolBench~\cite{toolllm} focuses on multi-step reasoning, and BFCL v4~\cite{bfcl} evaluates proficiency in multi-turn interactions. 
To assess domain-specific abilities, representative benchmarks such as WebArena~\cite{WebArena}, SWE-bench~\cite{swebench}, and OSWorld~\cite{osworld} evaluate agent performance in web navigation, software development, and operating system environments, respectively. However, these benchmarks operate in static environments, contrasting sharply with the dynamic and evolving nature of real-world settings. 
This discrepancy creates a critical gap in evaluating the adaptability of LLM agents to continuous environmental changes.

\section{Empirical Study of MCP Server Evolution }
\label{sec:empirical_study}
To validate the existence of real-world MCP server evolution and characterize its dynamics, we conducted a large-scale empirical study from two dimensions: the temporal dynamics of remotely deployed servers and the evolutionary history of publicly released versions.

\subsection{MCP Server Collection}
\label{subsec:data_collection}
Our data acquisition process comprises two parallel streams: remote server monitoring and released repository collection. 
For remote services, we used \texttt{Smithery}~\cite{smithery}, a mainstream MCP server hosting platform as our data source. We used functional keywords such as ``browser-automation'' to identify 1,869 repositories hosting remotely accessible MCP servers. Each service endpoint was subjected to continuous availability monitoring for three months, with weekly testing intervals.
For code repositories, we aggregated server names from \texttt{Smithery}, \texttt{GitHub}~\cite{github}, and \texttt{ModelScope}~\cite{modelscope} and traced their publication histories via the \texttt{NPM} package manager~\cite{npm_registry}. 
After validation through actual deployment, we curated a dataset of 515 independent MCP servers, containing 9,273 historical versions with 6,436 distinct tools. Detailed collection process is provided in Appendix \ref{app:collection}.


\subsection{Analysis of MCP Server Evolution}
\begin{figure}[t] 
    \centering
    \begin{minipage}{0.52\textwidth}
        \centering
        \includegraphics[width=\linewidth]{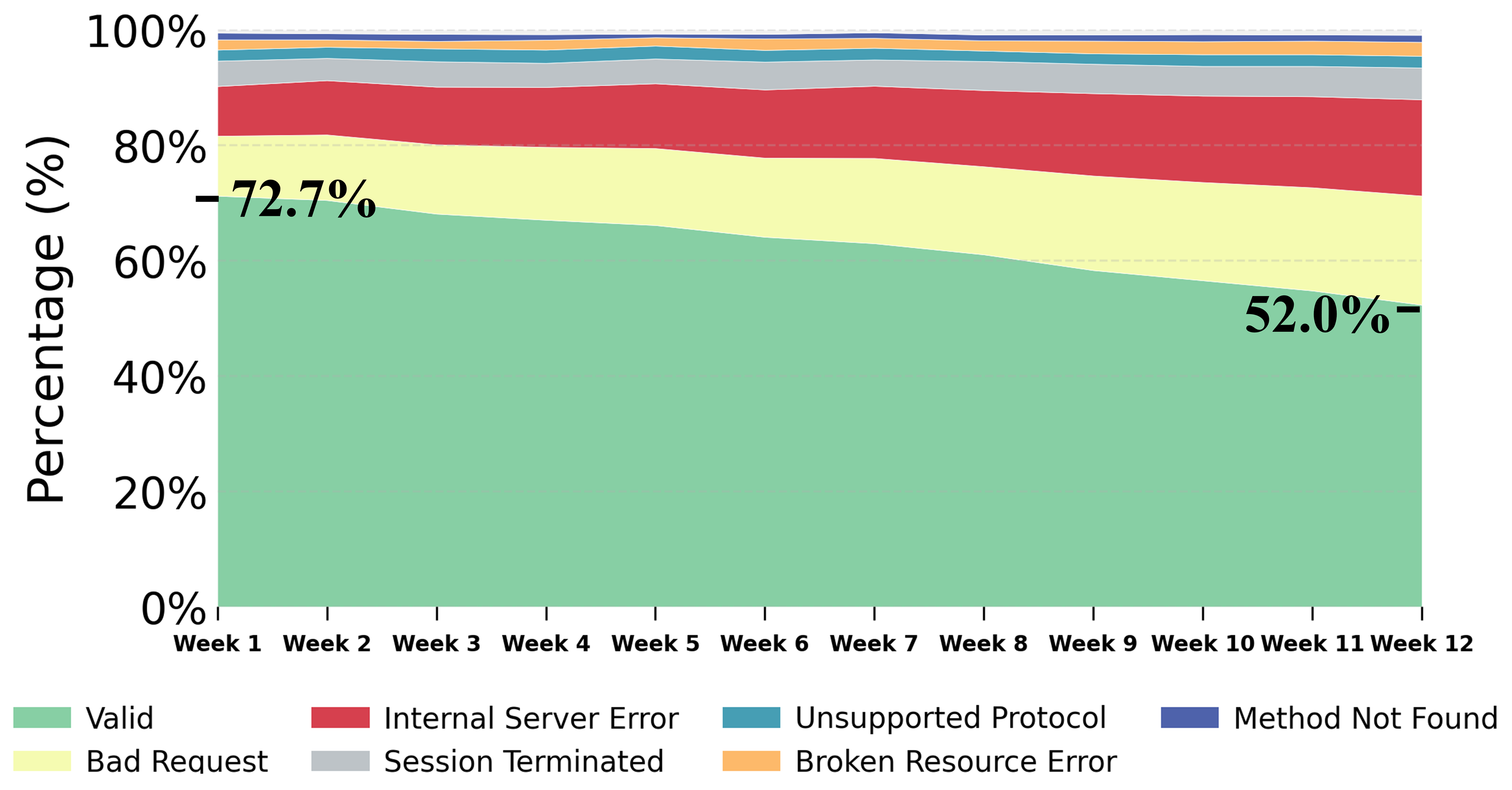}
        \caption{Availability of remote MCP servers.}
        \label{fig:ava}
    \end{minipage}
    \hfill 
    \begin{minipage}{0.47\textwidth}
        \centering
        \includegraphics[width=\linewidth]{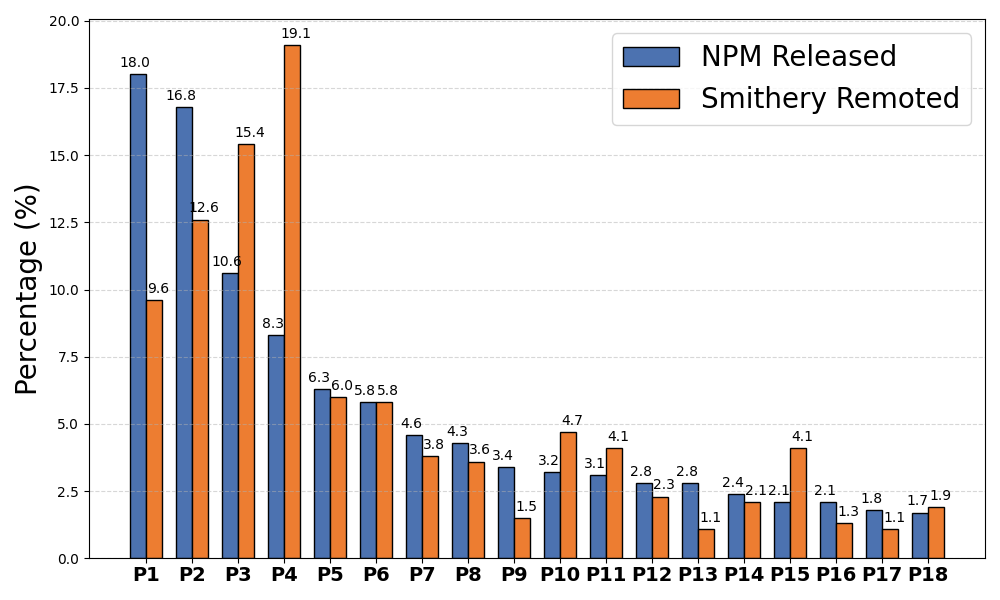}
        \caption{The distribution of evolution patterns.}
        \label{fig:pattern}
    \end{minipage}
    \vspace{-0.3em}
\end{figure}

\textbf{Remote Service Availability Decay.} Figure \ref{fig:ava} illustrates the validity of remote MCP servers decreased significantly from 72.7\% (week 1) to 52.0\% (week 12) over three months. 
This decline is primarily driven by \textit{Bad Request} (18.9\%) and \textit{Internal Server Error} (16.7\%), both attributable to deployment failures arising from compatibility conflicts in the evolution process.

\textbf{Growth of Toolset Complexity.} Following the release sequence, we analyzed the evolution trends of MCP servers in terms of average tool count, parameter count, and description length. As shown in Figure \ref{fig:trend}, all three metrics exhibit an overall upward trend across version iterations, showing an expansion in toolset scale and increasing structural complexity. 
Additionally, we find that 54.6\% of the initial tools were either modified (32.5\%) or deprecated (22.1\%) in the latest versions, reflecting the intense evolutionary activity of MCP servers.


\textbf{Analysis of Evolution Patterns.} 
We analyzed tool evolution across version iterations in NPM-published MCP servers and along the temporal timeline in Smithery-deployed servers.
Figure \ref{fig:pattern} presents the distribution of these evolution patterns, with types such as P1 (TOOL-DESC\_CHANGE) for tool description modifications and P2 (TOOL-ADD) for new tool additions; full definition of these patterns are available in Appendix \ref{app:pattern}. 
We observe two main characteristics: 1) \textbf{Consistency}, where the top six patterns (P1–P6) are identical across both data sources, suggesting that these evolution patterns are platform-independent and accurately reflect the prevailing evolution dynamics; and 2) \textbf{Divergence}, where Smithery-deployed servers prefer parameter-level adjustments (P3, P4), in contrast to the tool-level modifications (P1, P2) dominant in NPM repositories. This divergence likely stems from minimizing tool-level changes to ensure remote service continuity.

In summary, our findings reveal the current landscape of MCP server evolution and also highlight the critical need for benchmarks that evaluate LLM agents' adaptability in evolving environments.

\section{MCPEvol-Bench Formalization}
The agent task with environment feedback is formalized as a Partially Observable Markov Decision Process (POMDP). Each task in our benchmark is defined by a POMDP tuple $(\mathcal{U}, \mathcal{S}, \mathcal{A}, \mathcal{O}, T, R, \Sigma)$, where:  $\mathcal{U}$ denotes the task instruction space; $\mathcal{S}$ is the state space; $\mathcal{A}$ is the action space including both reasoning content and tool invocations; $\mathcal{O}$ is the observation space containing tool execution results and internal signals; $T: \mathcal{S} \times \mathcal{A} \rightarrow \mathcal{S} \times \mathcal{O}$ is the state transition function; $R: \mathcal{S} \rightarrow [0,1]$ is the reward function; $\Sigma = \{v_1, v_2, \dots, v_n\}$ is the set of available MCP servers.
Unlike most benchmarks~\cite{mcp_bench, livemcpbench, mcpeval} that rely on dynamic tool retrieval, we maintain fixed MCP server selection for each task to isolate and evaluate the impact of server evolution on agent workflows.
Each server $v_i \in \Sigma$ exposes a set of tools $\mathcal{T}_i$, defining the complete toolset $\mathcal{T} = \bigcup_{i=1}^n \mathcal{T}_i$. A structured tool invocation is written as $a_{\text{tool}} = \langle v_i, \texttt{tool\_name}, \texttt{parameters} \rangle$. The full action space is $\mathcal{A} = \mathcal{A}_{\text{reasoning}} \cup \mathcal{A}_{\text{tools}}$, and the observation space is $\mathcal{O} = \mathcal{O}_{\text{tools}} \cup \mathcal{O}_{\text{state}}$. 

\begin{algorithm}[htbp]
\caption{Multi-turn Tool Invocation and Observation}
\label{alg:multiturn}
\begin{algorithmic}[1]
\Require Task instruction $u$, maximum steps $T_{\text{max}}$, MCP server cluster $\Sigma$
\Ensure Final answer $\texttt{answer}$, execution trajectory $\texttt{trajectory}$

\Function{MULTITURNEXECUTE}{$u, T_{\text{max}}, \Sigma$}
    \State $\texttt{trajectory} \gets \{\}$, $s_0 \gets \textsc{Update}(u, \Phi(\Sigma))$ \Comment{Initialize initial trajectory and state}
    \For{$t = 0$ to $T_{\text{max}}$}
        \State $(\texttt{continue}_t, a_t) \gets \pi_{\text{}}(s_t)$ \Comment{Generate current tool invocation}
        \If{$\texttt{continue}_t = \texttt{False}$}
            \State $\texttt{answer} \gets a_t$ \Comment{Obtain the final answer from agent}
            \State \textbf{break} \Comment{Stop if agent signals termination}
        \EndIf
        \State $o_t \gets \Sigma(a_t)$ \Comment{Execute tools in MCP servers}
        \State $\texttt{trajectory} \gets \texttt{trajectory} \cup \{(a_t, o_t)\}$ \Comment{Log tool invocation and observation}
        \State $s_{t+1} \gets \textsc{Update}(s_t, o_t)$ \Comment{Update agent internal state}
    \EndFor
    \State \Return $(\texttt{answer}, \texttt{trajectory})$
\EndFunction
\end{algorithmic}
\end{algorithm}

For the workflow of the agent, we adopt a multi-round decision process~\cite{react}, as detailed in Algorithm \ref{alg:multiturn}. 
The agent initializes its execution trajectory and initial state $s_0$ with a server description function $\Phi: \Sigma \rightarrow \mathcal{D}$. The description space $\mathcal{D}$ includes essential metadata such as tool input schemas and semantic descriptions.
At each step $t$, the agent's policy $\pi$ generates an action $a_t$ that integrates reasoning with multiple tool calls, based on the current state $s_t$ (Line 4). 
Upon detecting a completion signal in $a_t$, the termination signal $\texttt{continue}_t$ is set \texttt{False}, and the reasoning content $a_t$ is returned as the final answer (Lines 5-8). 
Otherwise, the MCP server cluster $\Sigma$ executes the action $a_t$ to obtain the tool output as observation $o_t$ (Line 9). 
This observation is logged into $\texttt{trajectory}$, and the agent state $s_t$ is updated accordingly (Lines 10-11). The loop continues $T_\text{max}$ is reached or the task is completed. Finally, the full execution trajectory and answer are returned for evaluation. The prompt used for the agent execution can be found in Appendix \ref{app:prompt:eval}.

\section{MCPEvol-Bench Construction}
We present MCPEvol-Bench, a benchmark designed to evaluate the task-solving capabilities of LLM agents under dynamic toolset evolution, as illustrate in Figure \ref{fig:overview}. 
In this section, we first detail our task construction process.
Then, we describe how LLMs simulate human developers modifying and updating MCP toolsets.
Finally, through multi-round iterative evolution, we generate multi-version MCP servers, thereby establishing a realistic and dynamic environment for evaluating LLM agents.


\subsection{Task Synthesis.}
\textbf{MCP Servers.} The majority of MCP servers rely on proprietary API keys or third-party integrations, hindering plug-and-play deployment. To address this, we filtered our empirically collected MCP servers with an LLM-assisted process to exclude key-dependent instances. 
Besides accessibility, we ensure the toolset’s representativeness through structured curation and expert annotation.
The resulting collection comprises 123 MCP servers providing 1,272 tools and is taxonomically organized into nine functional categories: \textit{Knowledge} (13.8\%), \textit{Research} (7.3\%), \textit{Software Development} (19.5\%), \textit{UI Design} (6.5\%), \textit{Media \& Documentation} (16.3\%), \textit{Data \& Analytics} (8.1\%), \textit{Business \& Commerce} (10.6\%), \textit{AI \& Machine Learning} (14.6\%), \textit{Cloud \& Infrastructure} (3.3\%). To ensure reproducibility, we bundled the MCP servers into an NPM package, thereby providing a stable tooling environment. 

\textbf{Task Instructions.} Constructing high-quality benchmarks for tool-using agents requires transforming real-world MCP servers into realistic and solvable tasks. To this end, we propose an automated task synthesis pipeline. 
First, we employed random sampling to select 2–5 MCP servers from the same category, simulating cross-server scenarios.
Based on these servers, the LLM generated initial task instructions requiring multi-tool collaboration, without explicitly specifying the required servers and tools.
Following the methodology of MCP-Bench~\cite{mcp_bench}, each task underwent a rigorous two-dimensional quality assessment: 1) \textit{Solvability}, which evaluates whether the task can be successfully completed using the available tools; and 2) \textit{Practical Utility}, which determines whether the task addresses genuine user needs rather than contrived scenarios. 
Tasks failing to meet strict thresholds (Solvability $\geq$ 9.0/10; Utility $\geq$ 6.0/10) were discarded.
Finally, the remaining tasks were validated through LLM agent rollouts, where execution trajectories were assessed via both LLM judging and human review. Leveraging DeepSeek-Chat~\cite{deepseek-v3} as the task synthesis LLM, we constructed a final dataset of 201 high-quality tasks. On average, each sample involves 25.90 distinct tools within its task context, necessitating 4.37 tool invocations and the utilization of 3.76 servers for successful task completion. The prompt used for task synthesis can be found in Appendix \ref{app:prompt:construction}. Table \ref{tab:mcp_evolution_example} presents an example of a synthesized task and its associated MCP servers.

\begin{figure*}[t]
    \vspace{-0.5cm}
    \centering
    \includegraphics[width=0.8\linewidth]{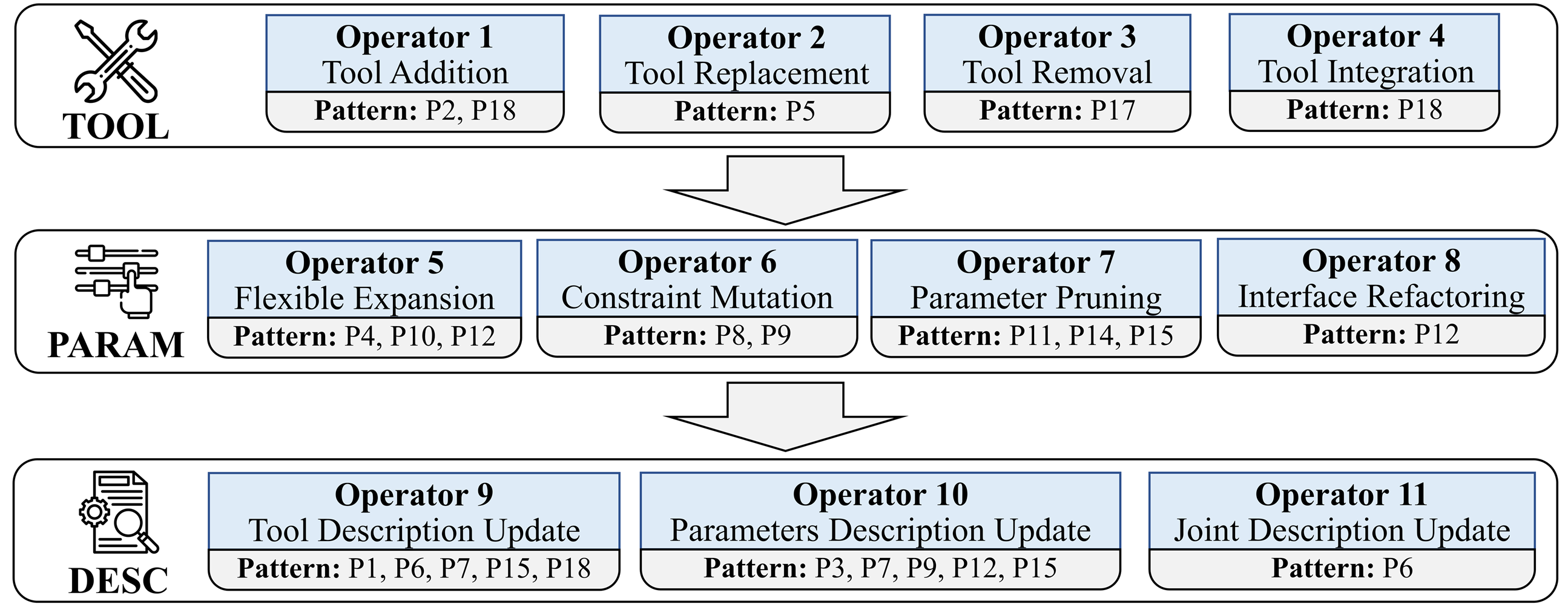} 
    \caption{
        The 11 evolution mutation operators are categorized into ``TOOL'', ``PARAM'', and ``DESC'' levels. These operators have a blue background, with associated evolution patterns (Figure \ref{fig:pattern}) in gray. 
        Arrows denote a hierarchical relationship, where higher-level mutations encompass lower-level ones.
    }
    \label{fig:operator}
    \vspace{-0.2cm}
\end{figure*}

\subsection{MCP Server Evolution}
\textbf{Evolution Mutation Operators.} 
Figure \ref{fig:operator} presents the 11 MCP server evolution operators derived from our empirical study, organized into three hierarchical levels: TOOL (tool), PARAM (parameter), and DESC (description). These operators are encapsulated as specialized prompts to guide LLMs in performing precise source code modifications for MCP tools. 
The evolution targets for each level are detailed below:
1) \textbf{``TOOL''}: This level includes four operators, namely \textit{Tool Addition}, \textit{Tool Replacement}, \textit{Tool Deletion}, and \textit{Tool Integration}. These operators implement functional changes by adding, removing, or substituting tools, and by simultaneously coordinating the functional logic of existing tools to ensure consistency.
2) \textbf{``PARAM''}: Comprising \textit{Flexible Expansion}, \textit{Constraint Mutation}, \textit{Parameter Pruning}, and \textit{Interface Refactoring}, these operators optimize parameter structures, constraints, data types, and required attributes, while preserving the core functionality of tools.
3) \textbf{``DESC''}: three operators function at this level, which are \textit{Tool Description Update}, \textit{Parameter Description Update}, and \textit{Joint Description Update}. They serve to refine natural language descriptions, eliminating ambiguity and clarifying functional boundaries to ensure efficient agent invocation.
The numerical labels (e.g., P1, P2) within each operator correspond to the evolution patterns identified in our empirical study (Figure \ref{fig:pattern}). 
Collectively, these operators cover all identified evolution patterns and can be composed to model more complex evolutionary scenarios, thereby showing the completeness of our approach.
Definitions and prompts of operators are detailed in the Appendix \ref{app:operator} and \ref{app:prompt:evoluation}.

\begin{wrapfigure}{r}{0.44\textwidth} 
    \centering
    \includegraphics[width=\linewidth]{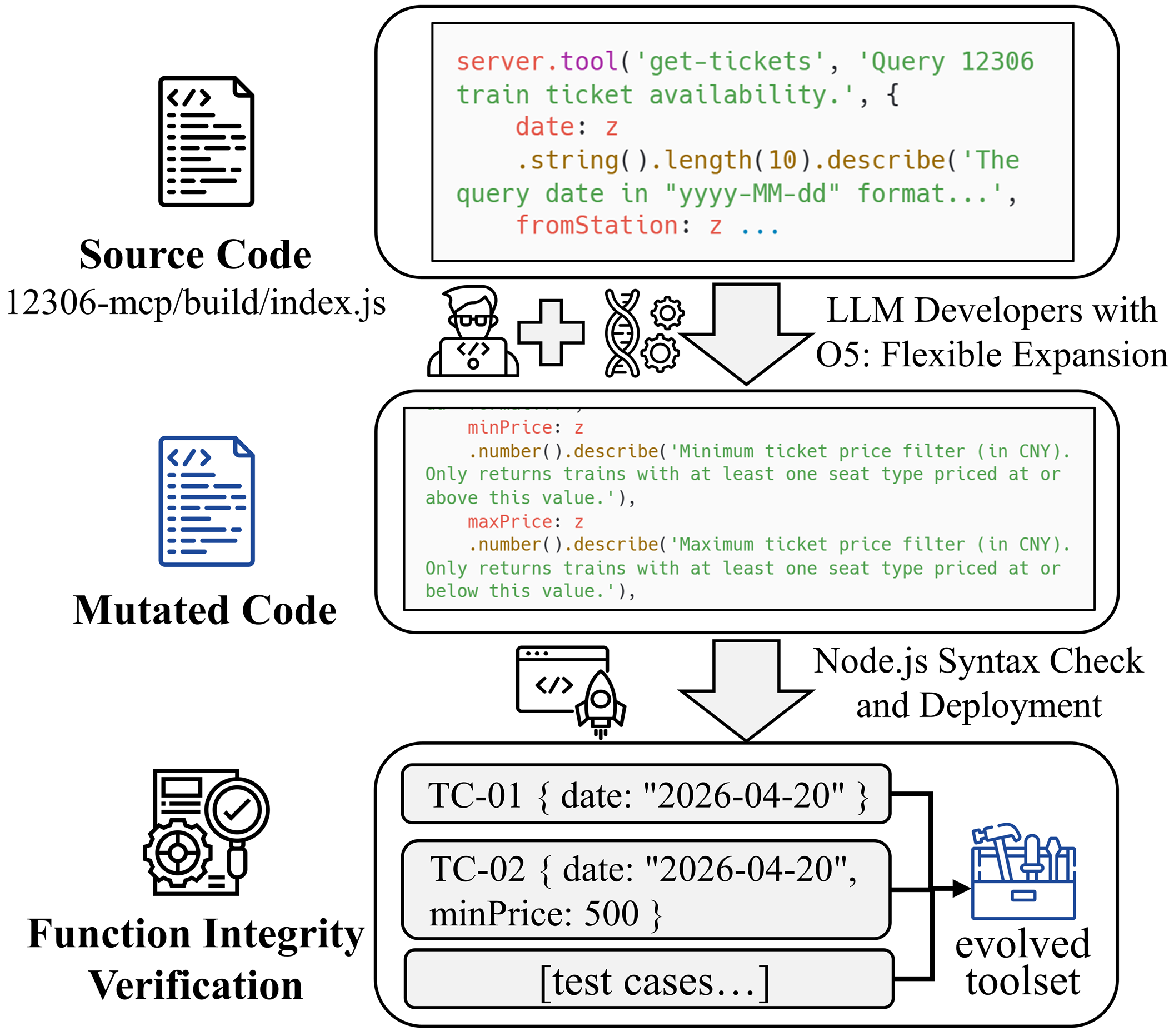}
    \caption{An example of LLM-Driven MCP Server Evolution.}
    \label{fig:example}
    \vspace{-0.2cm}
\end{wrapfigure}

\textbf{LLM-Driven Mutation Mechanism.} 
Real-world legacy MCP servers often quickly become non-functional due to invalid URLs and breaking changes in external dependencies or runtime environments. To address this, we propose an LLM-driven evolution mechanism that simulate realistic MCP server evolution by directly modifying source code repositories.
This mechanism follows a structured workflow, starting with Abstract Syntax Tree (AST)-based code anchoring (Appendix \ref{app:ast}), which utilizes the registration syntax (\texttt{server.tool}) to extract snippets of tool definitions and implementations, effectively filtering out irrelevant contextual noise.
Based on these anchored segments, an LLM (Claude-Opus-4-5) autonomously selects appropriate mutation operators and mutates the source code, preserving backward compatibility and original functionality.
Upon passing syntax checks, the mutated code repository is deployed as an MCP server for \textbf{Functional Integrity Validation}. In this phase, an LLM (DeepSeek-Chat) generates multiple test cases to perform comprehensive coverage testing on the mutated tools, validating their compliance with the mutation requirements and ensuring functional correctness.
This evolution process can be iterated over multiple rounds, with a single mutation operator applied per round, ultimately generating multi-version MCP servers.

Figure \ref{fig:example} illustrates this process, where the LLM applies Operator 5 (O5) to augment the \texttt{get-tickets} tool with \texttt{minPrice} and \texttt{maxPrice} parameters. This enhancement enables users to filter results by fare range and is subsequently verified through multi-case validation.

\section{Experiments}
\subsection{Experimental Setup}
\textbf{Benchmark Configurations.} To evaluate LLM agents’ task-solving capabilities under tool evolution, MCPEvol-Bench pairs each task with three MCP server configurations representing different evolution stages: \textbf{Early Stage} (original, non-evolved server), \textbf{Middle Stage} (evolved for 3 rounds), and \textbf{Late Stage} (evolved for 5 rounds). 
Prior studies~\cite{toolllm, livemcpbench} show that the primary source for agent failures is retrieving wrong task-required tools. 
To analyze the impact of server evolution, we bypass the dynamic retrieval process and employ a set of fixed candidate MCP servers to support task completion.

\textbf{Evaluated Models.} We evaluated 12 representative LLMs, including commercial models GPT-5.4~\cite{gpt54}, GPT-5.1~\cite{gpt5}, GPT-4o~\cite{gpt4}, Claude-Opus-4-6~\cite{claude_opus_46}, Claude-Sonnet-4-6~\cite{claude_sonnet_46}, and Gemini-2.5-Pro~\cite{gemini25}; long-reasoning models o3~\cite{o3} and Claude-Sonnet-4-5-thinking~\cite{claude_sonnet_45}; as well as open-source models Llama-3.3-70B~\cite{llama3}, Gemma-4-31B-it~\cite{gemma-4}, Qwen3.5-27B, and Qwen3.5-9B\cite{qwen35}.


\textbf{Evaluation Method and Metrics.}
We adopt the evaluation framework of MCP-Bench~\cite{mcp_bench}, leveraging the robustness of rubric-based LLM judges~\cite{judgebench, mllm, mcp_bench}.
LLM agents are assessed on a 1–10 scale with two dimensions: \textbf{Task Fulfillment} and \textbf{Planning Effectiveness}.
Evaluations are grounded solely in observable evidence from the task instruction, final answer, and execution trajectory. By default, the judge model used here is DeepSeek-Chat~\cite{deepseek-v3}.
To further evaluate the model's adaptability, we propose the \textbf{Evolutionary Competency Score} (ECS), which explicitly integrates performance with cross-version stability.
Specifically, for each task $u \in \mathcal{U}$, let $\{S_{u,i}^{\text{TF}}\}_{i=1}^N$ be the \textit{Task Fulfillment} scores at $i$-th version. The ECS is defined as:
\begin{equation}
    \text{ECS} = \frac{1}{|\mathcal{U}|} \sum_{u \in \mathcal{U}} \left( \mu(\{S_{u,i}^{\text{TF}}\}_{i=1}^{N}) - \sigma(\{S_{u,i}^{\text{TF}}\}_{i=1}^{N}) \right)
\end{equation}
where $N=3$, $\sigma$ and $\mu$ denote the standard deviation and mean, respectively.
The effectiveness of ECS metric and other evaluation details are provided in Appendix \ref{app:eval}.


\begin{table}[t]
    \centering
    \caption{Performance comparison of LLM agents across three evolution stages of MCP configuration, the early stage with the original servers, the middle stage after three evolution iterations, and the late stage after five evolution iterations.}
    \label{tab:main_results}
    \resizebox{\textwidth}{!}{%
    
    \begin{tabular}{l c c c c c c c}
        \toprule
        \multirow{2}{*}{\large\textbf{Model}} & 
        \multicolumn{2}{c}{\large\textbf{Early Stage}} & 
        \multicolumn{2}{c}{\large\textbf{Middle Stage}} & 
        \multicolumn{2}{c}{\large\textbf{Late Stage}} & 
        \multirow{2}{*}{\large\textbf{ECS} $\uparrow$} \\
        \cmidrule(lr){2-3} \cmidrule(lr){4-5} \cmidrule(lr){6-7}
        & \makecell{Task \\ Fulfillment} $\Big\uparrow$  & \makecell{Planning \\ Effectiveness} $\Big\uparrow$
        & \makecell{Task \\ Fulfillment} $\Big\uparrow$  & \makecell{Planning \\ Effectiveness} $\Big\uparrow$
        & \makecell{Task \\ Fulfillment} $\Big\uparrow$  & \makecell{Planning \\ Effectiveness} $\Big\uparrow$
        & \\
        \midrule
        Qwen3.5-9B     & 3.38 & 4.89  & 3.21 & 4.45 & 3.41 & 4.86 & 3.20 \\
        Llama-3.3-70B  & 3.84 & 3.56  & 3.88 & 3.74 & 3.93 & 3.94 & 3.56 \\
        Qwen3.5-27B    & 4.46 & 5.57  & 4.55  & 5.97 & 4.13 & 5.12 & 3.78 \\
        GPT-4o         & 5.24 & 2.85  & 5.03 & 2.74 & 4.43 & 2.78 & 3.80 \\
        Gemini-2.5-pro & 5.73 & 4.62  & 5.20 & 4.03 & 4.96 & 3.71 & 3.84 \\
        o3             & 5.79 & 3.79  & 5.23 & 3.21 & 5.07 & 3.10 & 4.24 \\
        GPT-5.1        & 6.28 & 4.35  & 5.94 & 3.43 & 5.32 & 3.50 & 4.32 \\
        Gemma-4-31B-it & 5.05 & 6.67  & 5.19 & \textbf{6.69} & 5.16 & \textbf{6.68} & 4.45 \\
        Claude-Sonnet-4-5-thinking   & 6.40 & 6.17 & 6.13 & 5.56 & 5.72 & 4.98 & 4.52 \\
        GPT-5.4        & \textbf{7.23} & 5.71  & \textbf{6.74} & 5.52 & 6.24 & 3.87 & 5.20 \\
   Claude-Sonnet-4-6   & 7.22 & \textbf{6.76}  & 6.61 & 6.42 & 6.18 & 6.17 & 5.22 \\
     Claude-Opus-4-6   & 7.15 & 6.70  & 6.93 & 6.18 & \textbf{6.77}& 6.03 & \textbf{6.09} \\
        \bottomrule
    \end{tabular}%
    }
    \vspace{-0.1cm}
\end{table}

\subsection{Main Results}
Table \ref{tab:main_results} reports the task performance of LLM agents across three MCP server evolution stages.

\textbf{The evolution of MCP servers induces substantial performance degradation and reduced execution efficiency across advanced LLMs.}
For instance, the task fulfillment scores for GPT-5.4 and Claude-Sonnet-4-6 exhibited a consistent decline from 7.23 and 7.22 in the early stage to 6.24 and 6.18  in the late stage, corresponding to degradation of 13.7\% and 14.4\%, respectively.
Concurrently, the execution efficiency of LLM agents decreases universally; notably, Claude outperformed the GPT series, which suffered a sharp drop in planning effectiveness, as evidenced by the fall in GPT-5.4’s score from the initial 5.71 to 3.87.

\textbf{Cross-version performance stability is crucial for the reliability of LLM agents.}
Although Claude-Opus-4-6, Claude-Sonnet-4-6, and GPT-5.4 exhibited comparable scores on the original server, only Claude-Opus maintained robust workflow execution during server evolution, achieving the highest ECS of 6.09. 
In contrast, the other two models showed significant performance degradation on the evolved MCP servers.
Similarly, while GPT-5.1 started with a higher task fulfillment score of 6.28 compared to Gemma-4-31B-it's 5.06, Gemma exhibited superior stability throughout subsequent evolution rounds, ultimately surpassing GPT-5.1 with an ECS of 4.45 versus 4.32.

\textbf{Models with limited capabilities are insensitive to environmental changes.}
For instance, Qwen3.5-9B and Llama-3.3-70B exhibited consistently low task fulfillment scores (<4.0), reflecting a failure to execute expected workflows. 
The ECS metric effectively captures this by penalizing low performance despite minimal variance, clearly separating true adaptability from mere invariance.

\subsection{In-depth Analysis}
\label{section:analy}
Our analysis focus on three frontier models: Claude-Opus-4-6, Claude-Sonnet-4-6, and GPT-5.4. 

\begin{figure*}[t] 
    \centering
    \begin{minipage}{0.51\textwidth}
        \centering
        \includegraphics[width=\linewidth]{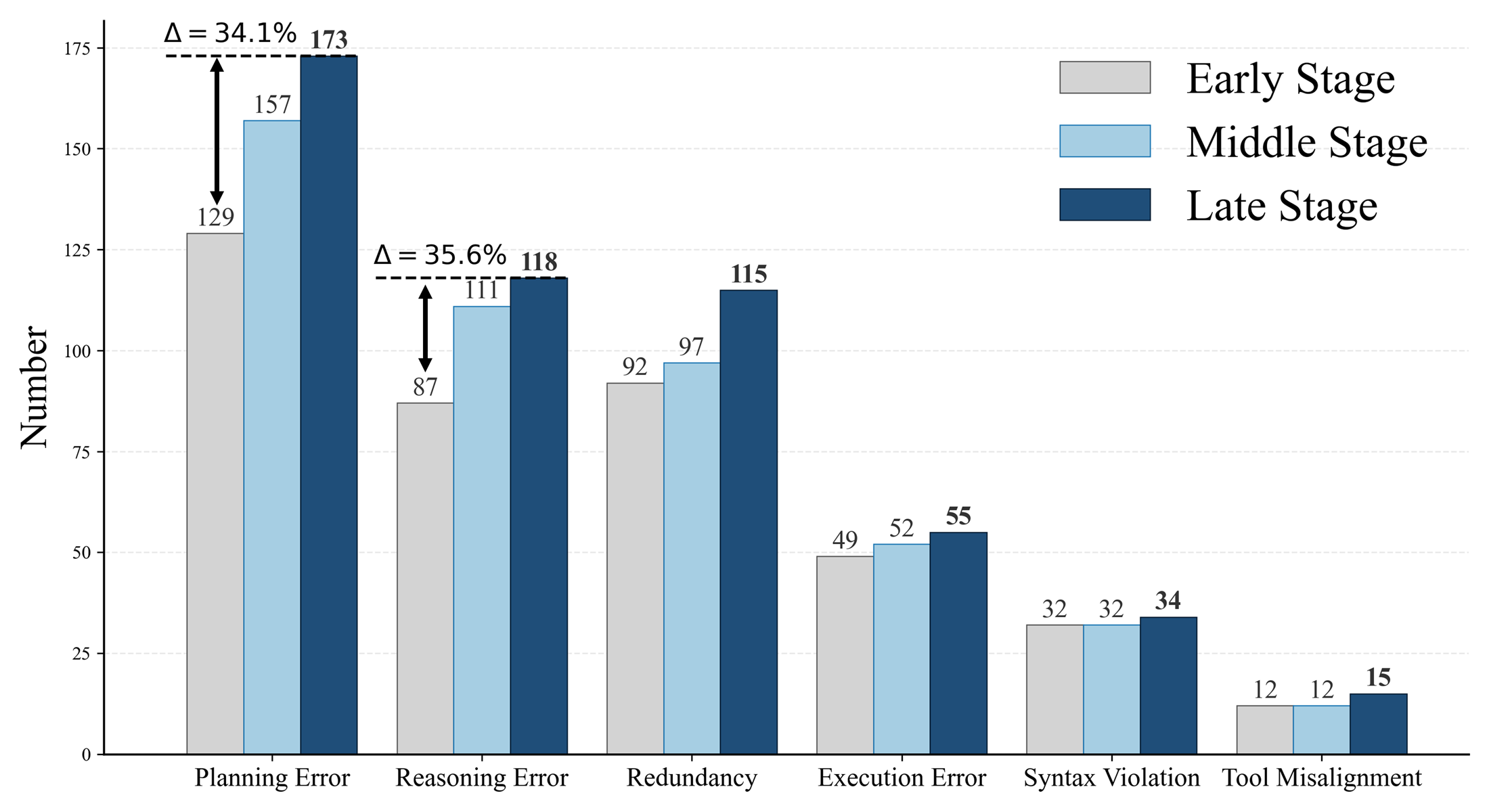}
        \par\vspace{0.05em} 
        (a) Error Distribution.
    \end{minipage}
    \hfill 
    \begin{minipage}{0.465\textwidth}
        \centering
        \includegraphics[width=\linewidth]{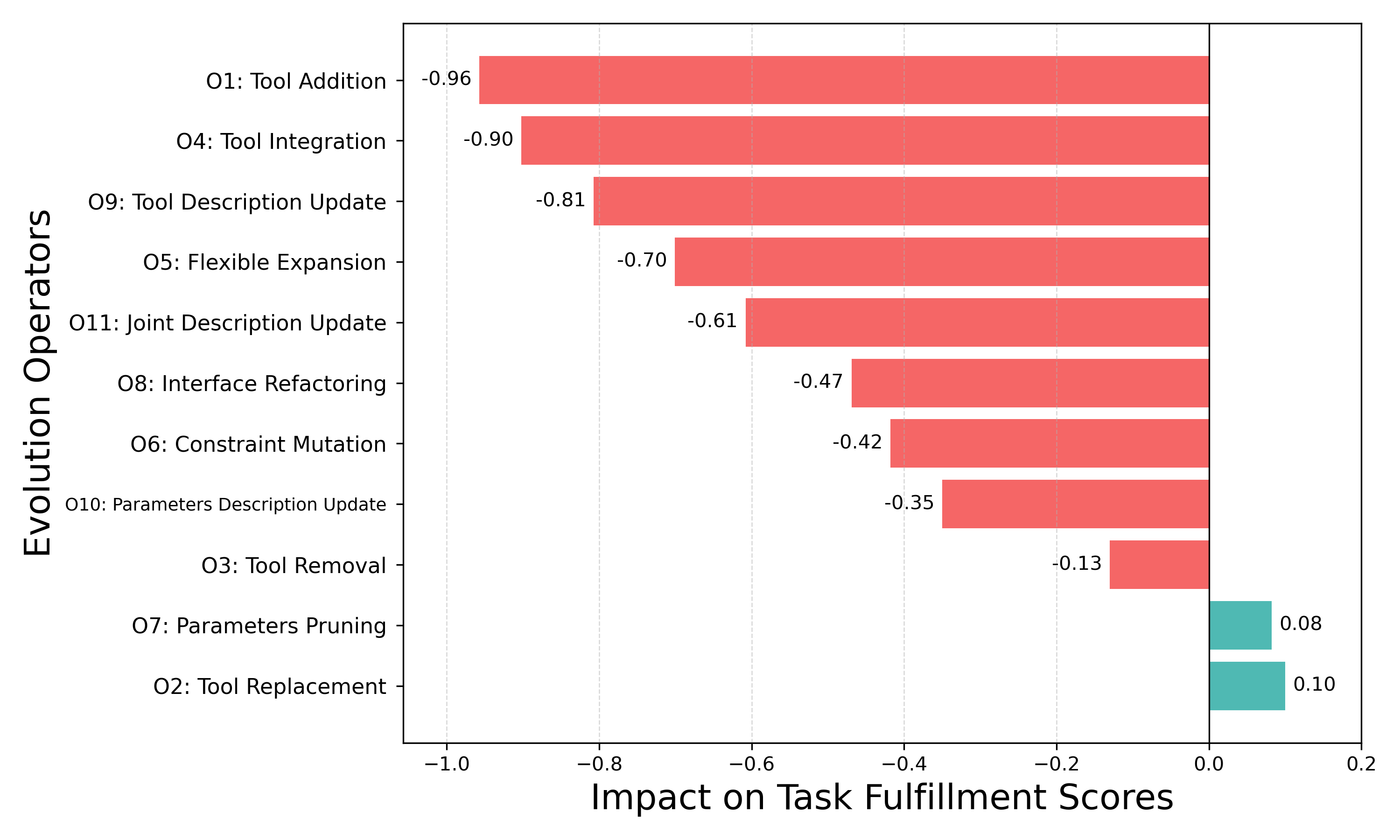}
        \par\vspace{0.05em} 
        (b) Evolution Operator Impact.
    \end{minipage}
    \caption{Diagnostic Analysis of MCP Server Dynamic Evolution. (a) Universal increase in error types across evolution stages. (b) The impact of each evolution operator on task fulfillment metric.}
    \label{fig:diagnostic}
    \vspace{-0.2cm}
\end{figure*}

\textbf{Impact of Evolution on Agent Behavior.}
We assessed the impact of evolution on agent behavior by analyzing the error distribution in task trajectories.
Building upon prior work~\cite{toolllm, agentrx, trajad}, we identified six primary categories of failures: 1) \textit{Syntax Violation}: violations of the tool-calling protocol, such as invalid parameter names or schema violations; 2) \textit{Tool Misalignment}: selection of inappropriate tools that fail to align with the user’s intent or current sub-task; 3) \textit{Execution Error}: provision of semantically incorrect parameter values; 4) \textit{Planning Error}: errors in multi-step workflow orchestration, such as missing preconditions, infinite loops, or premature termination;  5) \textit{Reasoning Error}: failure to correctly interpret or react to tool outputs; and 6) \textit{Redundancy}: inefficient execution paths by repeated identical calls or unnecessary actions. 
Figure \ref{fig:diagnostic} (a) shows the distribution of error types on MCP servers at various stages of evolution.
It is observed that MCP server evolution leads to increased erroneous behaviors in LLM agents.
Specifically, the increase in agent errors was concentrated in reasoning and planning errors, which rely heavily on context understanding, rising by 34.1\% and 35.6\%, respectively. 
Conversely, errors from tool misalignment and syntax violations did not increase significantly. This suggests that the context drift introduced by the evolution of MCP servers did not impair the model’s intrinsic tool-calling capability, but rather disrupted its previously stable workflow, causing LLM agents to generate many redundant errors through trial-and-error.

\textbf{Impact of Evolution Mutation Operators.}
We recorded the evolution operators selected by LLM developers for MCP servers over multiple rounds of evolution.
By comparing task fulfillment scores of the same task at different evolution stages, we analyzed each operator’s impact on task completion.
Figure \ref{fig:diagnostic} (b) presents the operator-specific contributions, revealing three key insights:
1) Evolution operations predominantly degrade LLM agents' performance. The majority of operators (9 out of 11) exhibited a negative impact, showing that the evolution process tends to disrupt established workflows.
2) Modifying tool descriptions and adding new tools resulted in the most negative effect. Operators such as O1 (Tool Addition, -0.96), O4 (Tool Integration, -0.90), and O9 (Tool Description Update, -0.81) substantially lowered scores. Such extensive changes often disrupt the agent’s tool orchestration logic, thereby hindering execution accuracy.
3) Subtractive operations are relatively neutral. Operators for removal or pruning had minimal negative or positive effects, as seen in O2 (Tool Replacement, +0.10), O7 (Parameters Pruning, +0.08), and O3 (Tool Removal, -0.13). 
When executing these updates, developers usually eliminate redundancies while preserving core MCP server functionalities, which reduces the workflow complexity and facilitates task completion.


\textbf{Experiments on real evolution and agent modules.}
We curated a subset of 50 historical versions of real-world MCP servers, covering 86 tasks, by excluding undeployable or non-functional instances.
\begin{wraptable}{r}{0.48\textwidth}
  \centering
  \centering
  \caption{Performance gains from agent modules.}
  \label{tab:agent}
  \renewcommand{\arraystretch}{0.85} 
\footnotesize
  \begin{tabular}{lcc}
    \toprule
    \textbf{Method} & \makecell{Task \\ Fulfillment} $\Big\uparrow$ & \makecell{Planning \\ Effectiveness} $\Big\uparrow$ \\
    \midrule
    Vanilla & 6.24 & 3.87 \\
    \quad + Reflection~\cite{reflexion} & 6.60 & 4.12 \\
    \quad + Plan~\cite{plan} & 6.52 & 4.23 \\
    \quad + Memory~\cite{awm} & \textbf{6.71} & \textbf{5.04} \\
    \bottomrule
  \end{tabular}
    \vspace{-0.1cm}
\end{wraptable}
Table \ref{tab:real_mcp_performance} shows apparent declines in task fulfillment across all models. 
This confirms that our simulated evolution-induced issues also persist in real-world scenarios.
Furthermore, we enhanced GPT-5.4 with agent modules, including reflection, planning, and memory, and tested them on the late stage of MCP Servers. Table ~\ref{tab:agent} demonstrates that these methods effectively alleviated performance drops due to tool evolution, outperforming the vanilla model in both evaluation metrics.

\textbf{Reliability of MCPEvol-Bench.}
We compared the cosine similarity of code change embeddings for a single mutation or version update under three settings: \textit{Evol vs. Real}, which measures the similarity between our simulated evolutions and real-world updates; \textit{Real vs. Real}, which compares consecutive real-world versions; and \textit{Random}, a baseline established by randomly pairing updates from different servers.
As shown in Table \ref{tab:code_similarity}, while \textit{Evol vs. Real} scored slightly lower than \textit{Real vs. Real} on the general BGE-M3 model, it surpassed \textit{Real vs. Real} on the code-specific CodeT5 and StarCoder2 models.
This demonstrates that our method effectively simulates human developer behavior.
Furthermore, as shown in Table \ref{tab:evaluator_correlation}, the high consistency between human experts and LLM-judges in ranking trajectories across evolution stages validated the effectiveness of our assessment.
\begin{table}[htbp]
    \vspace{-0.5cm}
    \begin{minipage}{0.43\textwidth}
        \centering
        \caption{\footnotesize Performance drop on historical MCP server.}
      \label{tab:real_mcp_performance}
      \footnotesize
      \begin{tabular}{lcc}
        \toprule
        \textbf{Model} & Current & Historical \\
        \midrule
        GPT-5.4 & 7.56 & 6.63 \textcolor{red}{\tiny-12.3\%} \\
        Claude-Sonnet-4-6 & 7.33 & 6.47 \textcolor{red}{\tiny-11.7\%} \\
        Claude-Opus-4-6 & 7.40 & 7.10 \textcolor{red}{\tiny-4.1\%} \\
        \bottomrule
      \end{tabular}
    \end{minipage}
    \hfill
    \begin{minipage}{0.57\textwidth}
    \centering
    \caption{Semantic similarity of code changes.}
    \label{tab:code_similarity}
    \footnotesize
    \begin{tabular}{lccc}
        \toprule
        \textbf{Setting} & BGE-M3~\cite{bge-m3} & CodeT5~\cite{codet5} & StarCoder2~\cite{starcoder} \\ \midrule
        Random  & 0.42 & 0.32 & 0.30 \\
        Real vs. Real  & \textbf{0.71} & 0.46 & 0.45 \\
        Evol vs. Real & 0.63 & \textbf{0.52} & \textbf{0.53} \\
        \bottomrule
    \end{tabular}
    \end{minipage}
    \vspace{-0.5cm}
\end{table}


\begin{figure*}[t]
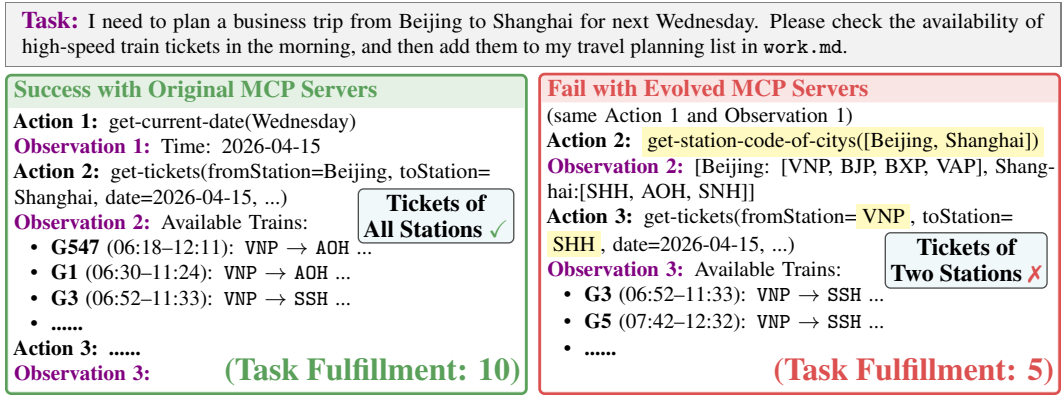

\setlength{\belowcaptionskip}{0pt}
\centering

\begin{tcolorbox}[
  colback=gray!10!white,
  colframe=black!50,
  boxsep=2pt,
  left=4pt, right=4pt, top=1pt, bottom=1pt,
  arc=2pt,
  boxrule=0.5pt,
  sharp corners,
  fontupper=\footnotesize,
  width=\linewidth 
]
\noindent
\textbf{\textcolor{obsPurple}{\small{Task:}}} I need to plan a business trip from Beijing to Shanghai for next Wednesday.
Please check the availability of high-speed train tickets in the morning, and then add them to my travel planning list in \texttt{work.md}. 
\end{tcolorbox}
\vspace{-0.15cm}
\begin{minipage}[t]{0.495\linewidth}
\sethlcolor{passgreen} 
\begin{tcolorbox}[
  enhanced,
  colback=white,
  colframe=passgreen!15!white,        
  borderline={1.0pt}{0pt}{passgreen}, 
  boxsep=1pt, left=1pt, right=1pt, top=1pt, bottom=1pt,
  arc=1pt,
  fontupper=\footnotesize,
  halign upper=left, 
  title={\textcolor{passgreen}{\bfseries \small Success with Original MCP Servers}}, 
  overlay={
      \node[
          anchor=south east, 
          xshift=-5pt, yshift=57pt, 
          draw=gray!60!black,       
          fill=unfaithfulbg,            
          line width=0.5pt,         
          rounded corners=2pt,      
          inner sep=2pt,            
          text=black,            
          font=\small\bfseries,
          align=center
      ] 
      at (frame.south east) {Tickets of \\ All Stations \textcolor{passgreen}{\checkmark} };
  }
]

\textbf{Action 1:} get-current-date(Wednesday) \\
\textcolor{obsPurple}{\textbf{Observation 1:}} Time: 2026-04-15 \\
\textbf{Action 2:} get-tickets(fromStation=Beijing, toStation= Shanghai, date=2026-04-15,  ...) \\
\textcolor{obsPurple}{\textbf{Observation 2:}} Available Trains: \\
\begin{itemize}[leftmargin=10pt, labelsep=1pt, nosep, itemsep=0pt]
    \item \enspace \textbf{G547} (06:18--12:11): \texttt{VNP} $\to$ \texttt{AOH} ...
    \item \enspace \textbf{G1} (06:30--11:24): \texttt{VNP} $\to$ \texttt{AOH} ...
    \item \enspace \textbf{G3} (06:52--11:33): \texttt{VNP} $\to$ \texttt{SSH} ...
    \item \enspace \textbf{......}
\end{itemize}
\textbf{Action 3:} \textbf{......} \\
\textcolor{obsPurple}{\textbf{Observation 3:}} \hfill \raggedleft \textcolor{passgreen}{\large\textbf{(Task Fulfillment: 10)}} 
\end{tcolorbox}
\end{minipage}
\hfill
\begin{minipage}[t]{0.495\linewidth}
\sethlcolor{failred}
\begin{tcolorbox}[
  enhanced,
  colback=white,
  colframe=failred!15!white,        
  borderline={1.0pt}{0pt}{failred}, 
  boxsep=1pt, left=1pt, right=1pt, top=1pt, bottom=1pt,
  arc=1pt,
  fontupper=\footnotesize,
  halign upper=left, 
  title={\textcolor{failred}{\bfseries \small Fail with Evolved MCP Servers}},
    overlay={
      \node[
          anchor=south east, 
          xshift=-5pt, yshift=40pt, 
          draw=gray!60!black,       
          fill=unfaithfulbg,             
          line width=0.5pt,         
          rounded corners=2pt,      
          inner sep=2pt,            
          text=black,            
          font=\small\bfseries,
          align=center
      ] 
      at (frame.south east) {Tickets of \\ Two Stations \textcolor{failred}{\ding{55}}};
  }
]
 (same Action 1 and Observation 1) \\
\textbf{Action 2:} \colorbox{yellow!30}{get-station-code-of-citys([Beijing, Shanghai])} \\
\textcolor{obsPurple}{\textbf{Observation 2:}} [Beijing: [VNP, BJP, BXP, VAP], Shanghai:[SHH, AOH, SNH]] \\
    \textbf{Action 3:} get-tickets(fromStation=\colorbox{yellow!30}{VNP}, toStation= \colorbox{yellow!30}{SHH}, date=2026-04-15,  ...) \\
\textcolor{obsPurple}{\textbf{Observation 3:}} Available Trains: \\
\begin{itemize}[leftmargin=10pt, labelsep=1pt, nosep, itemsep=0.6pt]
    \item \enspace \textbf{G3} (06:52--11:33): \texttt{VNP} $\to$ \texttt{SSH} ...
    \item \enspace \textbf{G5} (07:42--12:32): \texttt{VNP} $\to$ \texttt{SSH} ...
    \item \enspace \textbf{......}
\end{itemize}
 \hfill  \raggedleft \textcolor{failred}{\large\textbf{(Task Fulfillment: 5)}} 
\end{tcolorbox}
\end{minipage}
\caption{Case Study: Impact of MCP Server Evolution. Table \ref{tab:mcp_evolution_example} lists the original and evolved toolset configurations. Areas highlighted in \colorbox{yellow!30}{yellow} show changes compared to the original workflow.}
\label{fig:case}
\vspace{-0.1cm}
\end{figure*}

\subsection{Case Study}
Figure \ref{fig:case} present a case study on workflow degradation from server evolution. 
The original workflow \textcolor{passgreen}{(green)} achieved a perfect score by directly calling \texttt{get-tickets} for all train tickets data between cities. 
Evolved parameter description in tool \texttt{get-tickets} (Table \ref{tab:mcp_evolution_example}) indicate first to retrieve station codes by calling \texttt{get-station-code-of-citys}.
This perturbation resulted in the evolved workflow \textcolor{failred}{(red)} only querying tickets for two stations, \texttt{VNP} and \texttt{SHH}, causing the score to degrade to 5.

\section{Conclusion}
In this paper, we introduced \textbf{MCPEvol-Bench}, a novel benchmark for evaluating the task-solving capabilities of LLM agents under dynamic toolset evolution. This benchmark bridges the gap in existing static evaluations that fail to capture agent adaptability in evolving tool environments.
Guided by our empirical study on MCP server evolution, we developed 11 mutation operators enabling LLMs to automatically and iteratively modify tools within 123 real-world MCP servers, thereby simulating realistic server evolution. 
Experiments on 12 state-of-the-art LLMs reveal that even frontier models, such as GPT-5.4 and Claude-Sonnet-4-6 suffer significant performance degradation, highlighting the critical need for LLM agents capable of constructing stable workflows in dynamic environments.

\newpage
{
\small
\bibliography{ref}
\bibliographystyle{unsrt}
}

\newpage
\newpage
\appendix
\definecolor{HeaderDarkGray}{RGB}{80, 80, 80} 
\definecolor{DeepBlackGray}{RGB}{40, 40, 40} 

\lstdefinelanguage{JavaScript}{
    keywords={break, case, catch, continue, debugger, default, delete, do, else, finally, for, function, if, in, instanceof, new, return, switch, this, throw, try, typeof, var, void, while, with, class, const, let, export, import, extends, super, from, as, async, await, yield},
    keywordstyle=\color{blue}\bfseries,
    ndkeywords={class, export, boolean, throw, implements, import, this},
    ndkeywordstyle=\color{darkgray}\bfseries,
    identifierstyle=\color{black},
    sensitive=false,
    comment=[l]{//},
    morecomment=[s]{/*}{*/},
    commentstyle=\color{green!60!black}\ttfamily,
    stringstyle=\color{red}\ttfamily,
    morestring=[b]',
    morestring=[b]"
}

\lstset{
    language=JavaScript,       
    basicstyle=\ttfamily\small,
    numbers=none,
    numberstyle=\tiny\color{gray},
    stepnumber=1,
    numbersep=5pt,
    backgroundcolor=\color{gray!10},
    showspaces=false,
    showstringspaces=false,
    showtabs=false,
    frame=single,
    tabsize=2,
    captionpos=b,
    breaklines=true,
    breakatwhitespace=false,
    title=\lstname
}

\section*{Appendix}

\section{Impact Statement}
\label{app:impact}
MCPEvol-Bench highlights the vulnerability of LLM agents to evolving toolsets. By exposing how performance drops when MCP servers change (e.g., updated parameters or descriptions), our work drives the development of more adaptive and robust agents. This paves the way for more stable automation workflows, reducing errors and boosting productivity in software development and daily tasks. Furthermore, by focusing on the open-source Model Context Protocol (MCP), we support a more open and interoperable AI ecosystem. However, as agents integrate more seamlessly with external tools, vulnerabilities or misleading inputs could lead to the unintended leakage of sensitive data (such as financial records or codebases). Additionally, frequent interface changes may create security blind spots that bypass traditional static checks, requiring greater attention to data privacy and security in dynamic environments.

\section{Safeguarding Statement}
\label{app:safe}
MCPEvol-Bench instructs LLM agents to operate within isolated directories, though potential risks of harmful operations on the host system remain. We mitigate this by enforcing strict file permissions to prevent unauthorized modification or deletion of critical files during evaluation. None of the 123 MCP servers included rely on proprietary API keys, ensuring no personal information is exposed. Furthermore, evolution is simulated via LLM-driven code mutations on locally stored NPM packages within a closed, sandboxed environment, rather than interacting with live production services. This eliminates the risk of disrupting real-world software services. These measures ensure the benchmark can be safely deployed for research without posing security or privacy threats.

\section{Limitation and Future Work}
\label{app:limit}
The MCP servers included in our benchmark are constrained. To ensure plug-and-play deployment, we excluded instances requiring external API key authorization. These servers usually have more frequent updates due to maintenance by dedicated companies or institutions. However, this exclusion does not compromise the validity of our benchmark or the generalizability of our conclusions. Specifically, the observed inadequacy of current models in adapting to dynamically evolving tool-using scenarios. In future work, we plan to incorporate additional officially maintained MCP servers with greater practical utility and functionality, such as those for Hugging Face and Google Maps.

\section{Details of Experiments}
\label{app:exp}
\subsection{Effectiveness of Evaluation Metrics}
\label{app:eval}
In our experiments, we adopted the evaluation metrics \textbf{Task Fulfillment} and \textbf{Planning Effectiveness} from MCP-Bench~\cite{mcp_bench}. Both metrics are evaluated using rubric-based LLM judges, with prompt shuffling employed to ensure the robustness of assessments (prompts detailed in \ref{app:prompt:eval}). 
The main experimental results in Table \ref{tab:main_results} are averaged over five independent evaluations. The corresponding  variances are reported in Table \ref{tab:variance_results}. The observed low variance (all $< 0.2$) shows the stability and effectiveness of these two metrics.

\begin{table}[htbp]
    \centering
    \caption{Variance ($\sigma^2$) of performance metrics across three evolution stages. }
    \label{tab:variance_results}
    \resizebox{\textwidth}{!}{%
    \begin{tabular}{l c c c c c c}
        \toprule
        \multirow{2}{*}{\large\textbf{Model}} & 
        \multicolumn{2}{c}{\large\textbf{Early Stage}} & 
        \multicolumn{2}{c}{\large\textbf{Middle Stage}} & 
        \multicolumn{2}{c}{\large\textbf{Late Stage}} \\
        \cmidrule(lr){2-3} \cmidrule(lr){4-5} \cmidrule(lr){6-7}
        & \makecell{Task \\ Fulfillment} $\sigma^2$  & \makecell{Planning \\ Effectiveness} $\sigma^2$
        & \makecell{Task \\ Fulfillment} $\sigma^2$  & \makecell{Planning \\ Effectiveness} $\sigma^2$
        & \makecell{Task \\ Fulfillment} $\sigma^2$  & \makecell{Planning \\ Effectiveness} $\sigma^2$ \\
        \midrule
        Qwen3.5-9B     & 0.04 & 0.12  & 0.08 & 0.03 & 0.11 & 0.07 \\
        Llama-3.3-70B  & 0.15 & 0.02  & 0.09 & 0.14 & 0.05 & 0.18 \\
        Qwen3.5-27B    & 0.07 & 0.16  & 0.11 & 0.06 & 0.13 & 0.09 \\
        GPT-4o         & 0.03 & 0.11  & 0.17 & 0.04 & 0.08 & 0.15 \\
        Gemini-2.5-pro & 0.12 & 0.05  & 0.06 & 0.13 & 0.10 & 0.02 \\
        o3             & 0.09 & 0.14  & 0.03 & 0.10 & 0.16 & 0.07 \\
        GPT-5.1        & 0.18 & 0.08  & 0.12 & 0.05 & 0.04 & 0.11 \\
        Gemma-4-31B-it & 0.06 & 0.13  & 0.15 & 0.09 & 0.07 & 0.14 \\
        Claude-Sonnet-4-5-thinking   & 0.11 & 0.04 & 0.10 & 0.17 & 0.06 & 0.12 \\
        GPT-5.4        & 0.05 & 0.10  & 0.14 & 0.08 & 0.13 & 0.03 \\
   Claude-Sonnet-4-6   & 0.13 & 0.07  & 0.04 & 0.11 & 0.15 & 0.09 \\
     Claude-Opus-4-6   & 0.10 & 0.15  & 0.08 & 0.06 & 0.12 & 0.16 \\
        \bottomrule
    \end{tabular}%
    }
\end{table}

The proposed \textbf{Evolutionary Competency Score (ECS)} effectively evaluates agent adaptability during the evolution of MCP Server tools.
Let $\mathcal{U}$ denote the set of evaluation tasks. For each task $u \in \mathcal{U}$, let $\mathbf{S}_u = \{S_{u,1}^{\text{TF}}, \dots, S_{u,N}^{\text{TF}}\}$ be the sequence of \textit{Task Fulfillment} scores across $N$ MCP server versions. We define the mean performance $\mu_u$ and standard deviation $\sigma_u$ as:
\begin{equation}
    \mu_u = \frac{1}{N} \sum_{i=1}^{N} S_{u,i}^{\text{TF}}, \quad 
    \sigma_u = \sqrt{\frac{1}{N-1} \sum_{i=1}^{N} \left(S_{u,i}^{\text{TF}} - \mu_u\right)^2}.
\end{equation}
Although the Coefficient of Variation~\cite{coefficient} ($CV_u = \sigma_u / \mu_u$) is frequently used to measure stability, it is insufficient for assessing adaptability as it decouples stability from absolute performance magnitude (e.g., a low-performing stable model may have the same $CV_u$ as a high-performing one).
To construct a comprehensive adaptability metric, we formulate a utility function $J_u$ that maximizes mean performance $\mu_u$ while penalizing relative instability. 
By applying a linear penalty factor $k$ to $CV_u$, we obtain $J_u = \mu_u (1 - k \cdot CV_u)$. Substituting the definition of $CV_u$ yields $J_u = \mu_u - k \sigma_u$. 
This derivation establishes that penalizing relative volatility is mathematically equivalent to subtracting a scaled absolute standard deviation from the mean. The ECS simplifies this relationship by setting $k=1$:
\begin{equation}
    \text{ECS} = \frac{1}{|\mathcal{U}|} \sum_{u \in \mathcal{U}} (\mu_u - \sigma_u)
\end{equation}

The efficacy of ECS in assessing adaptability stems from its sensitivity to performance regression. In our benchmark ($N=3$), consider two models on task $u$:
\begin{itemize}[leftmargin=2em, itemsep=1pt, topsep=0pt, parsep=0pt]
    \item \textbf{Model A (Low adaptability):} Scores $\{0.9, 0.5, 0.9\}$ yield $\mu_A \approx 0.77$ and $\sigma_A \approx 0.23$, resulting in $\text{ECS}_A \approx 0.54$.
    \item \textbf{Model B (High adaptability):} Scores $\{0.8, 0.8, 0.8\}$ yield $\mu_B = 0.80$ and $\sigma_B = 0.00$, resulting in $\text{ECS}_B = 0.80$.
\end{itemize}
Despite similar mean performance, the single performance drop in Model A significantly penalizes its ECS, resulting in a score substantially lower than that of Model B.
For $N=3$ with integer scores in $[1, 10]$, the ECS ranges from $4-\sqrt{27} \approx -1.20$ to $10$. 

\subsection{Reliability Analysis of Benchmark}
In Section \ref{section:analy}, we compare the semantic similarity between code modifications generated by simulated evolution and those from real-world MCP server version updates. Results presented in Table \ref{tab:code_similarity} demonstrate the effectiveness of our evolution. Specifically, for real-world version updates (Real vs. Real), we analyze the code differences between consecutive versions (e.g., between v0.3.2 and v0.3.1, or v0.3.1 and v0.3.0). For encoder model BGE-M3~\cite{bge-m3}, we directly compute the cosine similarity of word embedding vectors. For the other two decoder-only models, we extract the feature vectors from the final layer for cosine similarity calculation.

While human experts often struggle with absolute scoring, they excel at relative ranking. 
Leveraging this insight, we evaluated the alignment between LLM-based evaluations and human expert judgments.
We analyzed task trajectories generated by three advanced models: Claude-Sonnet-4-6, Claude-Opus-4-6, and GPT-5.4. 
Let $\mathcal{U}$ denote the set of all task instances. 
For each task $u \in \mathcal{U}$, we established a ground truth ranking vector $\mathbf{R}^{H}_u = [r^H_{u,early}, r^H_{u,middle}, r^H_{u,late}]$ by averaging the ranks assigned by three human experts, where $r \in [1, 2, 3]$.
For instance, if $\mathbf{R}^{H}_u = [1, 3, 2]$, it implies that the trajectory in early stage is ranked best, and that in late stage is second. 
Simultaneously, we employed three LLM evaluators (DeepSeek-Chat, GPT-4o-mini, and Gemini-2.5-pro) to compute \textit{Task Fulfillment} scores. These scores were converted into rank vectors $\mathbf{R}^{E}_u$.
The consistency between each LLM evaluator and the human ground truth was quantified using the average of spearman’s rank correlation coefficient $\bar\rho$~\cite{spearman}:
\begin{equation}
    \bar\rho = \sum_{u \in \mathcal{U}} \rho_u / |\mathcal{U}| = \sum_{u \in \mathcal{U}} (1 - \frac{6 \mathbf{d}_u^\top \mathbf{d}_u}{N(N^2 - 1)}) / |\mathcal{U}|
\end{equation}
where $\mathbf{d}_u = \mathbf{R}^{H}_u - \mathbf{R}^{E}_u$ is the difference in ranks in task $u$, and $N=3$. 

Table \ref{tab:evaluator_correlation} presents the correlation results. The high correlation scores across all evaluators demonstrate that our evaluations closely align with human perception of trajectory quality.

\begin{table}[htbp]
    \centering
    \caption{$\bar\rho$ between LLM evaluators and human expert rankings. The ranked items are task trajectories generated by Claude-Sonnet-4-6, Claude-Opus-4-6, and GPT-5.4.}
    \label{tab:evaluator_correlation}
\resizebox{0.7\textwidth}{!}{%
    \begin{tabular}{lccc}
    \toprule
    \multirow{2}{*}{\textbf{LLM Evaluator}} & \multicolumn{3}{c}{\textbf{Trajectory}} \\
    \cmidrule(lr){2-4} 
                                            & Claude-Sonnet-4-6 & Claude-Opus-4-6 & GPT-5.4 \\
    \midrule
    DeepSeek-Chat                           & 0.82              & 0.73              & 0.85    \\
    Gemini-2.5-pro                          & 0.80              & 0.75              & 0.85    \\
    GPT-4o-mini                             & 0.79              & 0.75              & 0.84    \\
    \bottomrule 
    \end{tabular}
}
\end{table}

\subsection{ Implementation Details}
\label{app:detail}
During the benchmark construction phase, we primarily employed the DeepSeek-Chat model for task synthesis, executing these tasks via the corresponding MCP Server to generate multi-tool invocation trajectories. For the evolution mutation of the MCP Server, we utilized the Claude-Opus-4-5 model, while test case generation continued to rely on DeepSeek-Chat. 

To ensure workflow stability and strict adherence to tool invocation formatting standards, the temperature parameter for all model evaluations was set to 0.2. 
The file cleanup script will automatically run after each evaluation to ensure that previous runs do not affect subsequent results.
All experiments were conducted on a server equipped with dual Intel(R) Xeon(R) Platinum 8358P CPUs @ 2.60GHz (totaling 64 logical cores), running CentOS 7.9. The software environment was built upon Python 3.12.12, Node.js 24.13.0, and npm 11.6.2.
The cost for a single evaluation is approximately 0.26\$, calculated using DeepSeek-V3.2 pricing rates (0.14\$ per million input tokens and 0.28\$ per million output tokens).

For the experiments involving Figure \ref{fig:diagnostic}(b), an MCP server may be subject to multiple operators simultaneously. We evenly distribute the resulting impact on the score among each operator to calculate their contribution scores.
\section{Details of the Empirical Study}
\label{app:empirical}
\subsection{Details of Empirical Data Collection}
\label{app:collection}
The empirical analysis draws on two primary sources: 1) deployment metadata of MCP servers periodically crawled from \textbf{Smithery} platform, and 2) versioned code repositories published on NPM.

We crawled remotely deployed MCP servers from Smithery platform by combining the keywords listed in Table \ref{tab:mcp_keywords} with the filter \texttt{is:remote}. After deduplication, we obtained 1,869 unique endpoints of MCP servers.

\begin{table}[b]
  \centering
  \caption{Functional keywords of MCP servers crawled from Smithery.}
  \label{tab:mcp_keywords}
  \resizebox{\textwidth}{!}{%
  \begin{tabular}{@{} l l l l @{}}
    \toprule
    \textbf{Keywords} & & &  \\
    \midrule
    Aggregators & Art \& Culture & Architecture \& Design & Browser Automation \\
    Bio & Cloud Platforms & Code Execution & Coding Agents \\
    Command Line & Communication & Customer Data Platforms & Databases \\
    Data Platforms & Delivery & Developer Tools & Data Science Tools \\
    Embedded System & File Systems & Finance \& FinTech & Gaming \\
    Knowledge \& Memory & Legal & Location Services & Marketing \\
    Monitoring & Multimedia Process & Research & Search \\
    Security & Social Media & Sports & Support \& Service Mgmt. \\
    Translation Services & Text-to-Speech & Travel \& Transportation & Version Control \\
    Workplace \& Productivity & Other Tools \& Integrations & & \\
    \bottomrule
  \end{tabular}
  }
\end{table}

To complement the Smithery dataset, we also aggregated MCP servers from three additional sources. 
\begin{itemize}[leftmargin=2em, itemsep=1pt, topsep=0pt, parsep=0pt]
    \item \textbf{ModelScope (12 keywords):} \texttt{browser-automation}, \texttt{search}, \texttt{communication}, \texttt{developer -tools}, \texttt{entertainment-and-media}, \texttt{file-systems}, \texttt{finance}, \texttt{knowledge-and-memory}, \texttt{location-services}, \texttt{art-and-culture}, \texttt{research-and-data}, \texttt{calendar-management}.
    \item \textbf{GitHub (2 repositories):} \url{https://github.com/punkpeye/awesome-mcp-servers}, \url{https://github.com/wong2/awesome-mcp-servers}.
    \item \textbf{NPM (14 keywords):} \texttt{mcp server}, \texttt{model context protocol}, \texttt{modelcontext -protocol}, \texttt{mcp-server}, \texttt{mcp\_server}, \texttt{create mcp server}, \texttt{mcp}, \texttt{mcp ai}, \texttt{mcp agent}, \texttt{@modelcontextprotocol}, \texttt{@modelcontextprotocol/server}, \texttt{mcp tools}, \texttt{mcp ai agent}, \texttt{mcp protocol implementation}.
\end{itemize}
We initially crawled 906, 1,040, and 1,912 servers from GitHub, Smithery, and ModelScope, respectively. By matching these names against the NPM registry, we download 1,461 repositories with their historical versions. After the removal of undeployable instances, 515 servers with 9273 historical versions remained for analysis.

\subsection{Analysis of Tool Evolution Stages}
\begin{figure}[htbp] 
    \centering
    \begin{minipage}{0.45\textwidth}
        \centering
        \includegraphics[width=\linewidth]{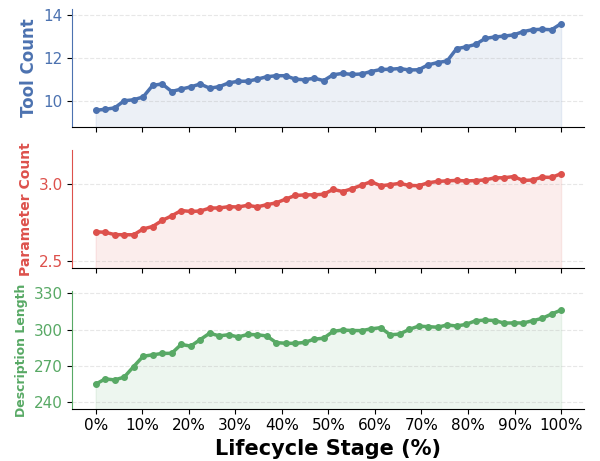}
        \caption{Growth of toolset complexity.}
        \label{fig:trend}
    \end{minipage}
    \hfill 
    \begin{minipage}{0.47\textwidth}
        \centering
        \includegraphics[width=\linewidth]{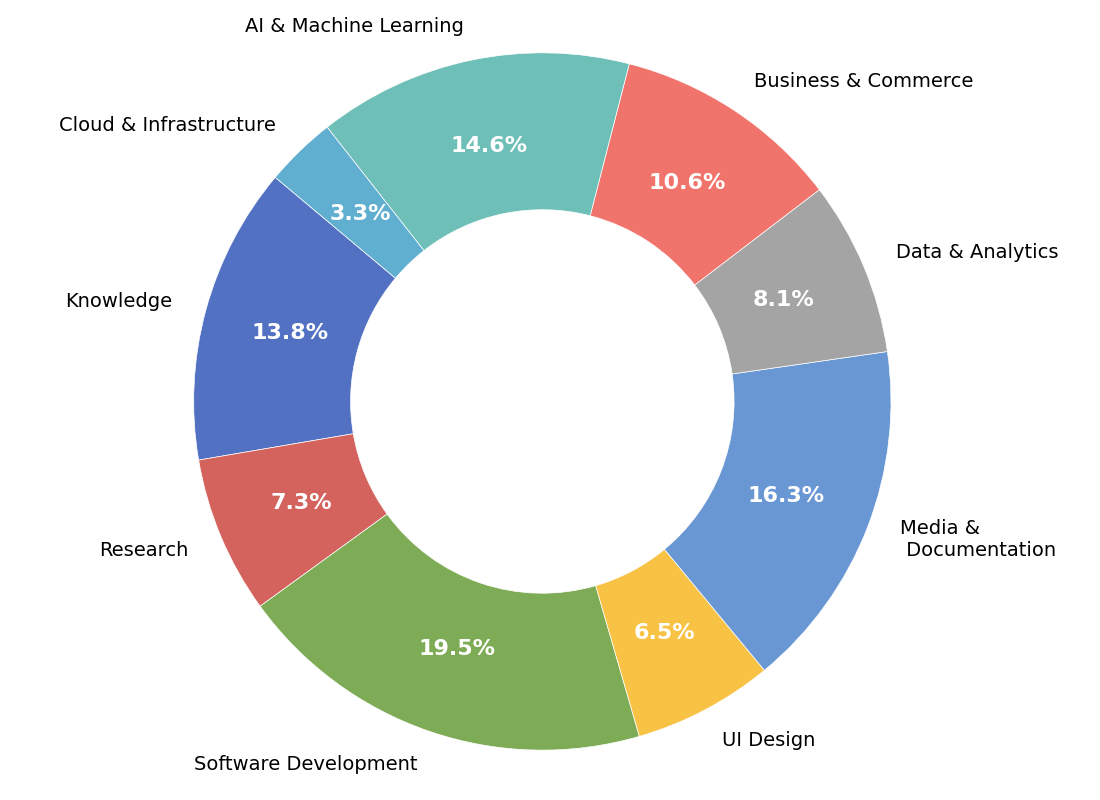}
        \caption{Function classification of MCP servers.}
        \label{fig:server_function}
    \end{minipage}
\end{figure}

To analyze tool evolution trends, we sorted the versions of the 515 collected MCP servers by release date and calculated their current evolution stage via normalization (e.g., version 8 out of 10 corresponds to 80\%). As shown in Figure \ref{fig:trend}, all three metrics exhibit significant growth.

\subsection{Analysis of Evolution Patterns}
\label{app:pattern}
The identified patterns capture the diverse ways in which MCP servers evolve over time, as shown in Table \ref{tab:evolution_patterns}. 
These changes range from simple semantic updates, such as modifying tool or parameter descriptions (P1, P3), to structural modifications like adding, removing, or replacing tools and parameters (P2, P4, P5, P10--P11, P14, P17). Additionally, servers often undergo complex composite changes (P6--P9, P12, P15--P16, P18) that simultaneously alter interface structures and semantic constraints. Understanding these patterns is crucial for developing agents that can robustly adapt to the dynamic nature of real-world tool environments.

\begin{table}[htbp]
  \centering
  \caption{Taxonomy of MCP Server Evolution Patterns}
  \label{tab:evolution_patterns}
  \resizebox{\linewidth}{!}{%
    \begin{tabular}{@{} l l l @{}}
      \toprule
      \textbf{ID} & \textbf{Pattern Name} & \textbf{Description} \\
      \midrule
      P1  & TOOL-DESC\_CHANGE & Update to the tool's natural language description. \\
      P2  & TOOL-ADD & Introduction of a new tool. \\
      P3  & PARAM-DESC\_CHANGE & Update to a specific parameter's description. \\
      P4  & PARAM-ADD(OPTIONAL) & Addition of a new optional parameter. \\
      P5  & TOOL-REPLACE & Replacement of an existing tool with a new implementation. \\
      P6  & TOOL-DESC\_CHANGE + PARAM-ADD(OPTIONAL) & Tool description update combined with optional parameter addition. \\
      P7  & TOOL-DESC\_CHANGE + PARAM-DESC\_CHANGE & Concurrent updates to both tool and parameter descriptions. \\
      P8  & PARAM-STRICT/RELEX & Modification of parameter validation constraints (strictness or relaxation). \\
      P9  & PARAM-STRICT/RELEX + PARAM-DESC\_CHANGED & Constraint modification accompanied by description update. \\
      P10 & PARAM-ADD(REQUIRED) & Addition of a new mandatory parameter. \\
      P11 & PARAM-REMOVE(OPTIONAL) & Removal of an existing optional parameter. \\
      P12 & PARAM-ADD(OPTIONAL) + PARAM-DESC\_CHANGE & Optional parameter addition with other parameters description update. \\
      P13 & PARAM-PROMOTE/DEMOTE & Change in parameter status between optional and required. \\
      P14 & PARAM-REMOVE(REQUIRED) & Removal of a previously mandatory parameter. \\
      P15 & PARAM-REMOVE(OPTIONAL) + PARAM-DESC\_CHANGE & Optional parameter removal with remaining parameters description updates. \\
      P16 & TOOL-DESC\_CHANGE + PARAM-ADD(OPTIONAL) & Tool description update combined with optional parameter addition. \\
      P17 & TOOL-REMOVE & Deprecation and removal of an existing tool. \\
      P18 & TOOL-ADD + TOOL-DESC\_CHANGE & New tool addition accompanied by other tool's description updates. \\
      \bottomrule
    \end{tabular}%
  }
\end{table}

\section{Details of Benchmark}
\label{app:benchmark}

\subsection{Statistics of MCPEvolBench}
\begin{figure}[htbp] 
    \centering
    \begin{minipage}{0.47\textwidth}
        \centering
        \includegraphics[width=\linewidth]{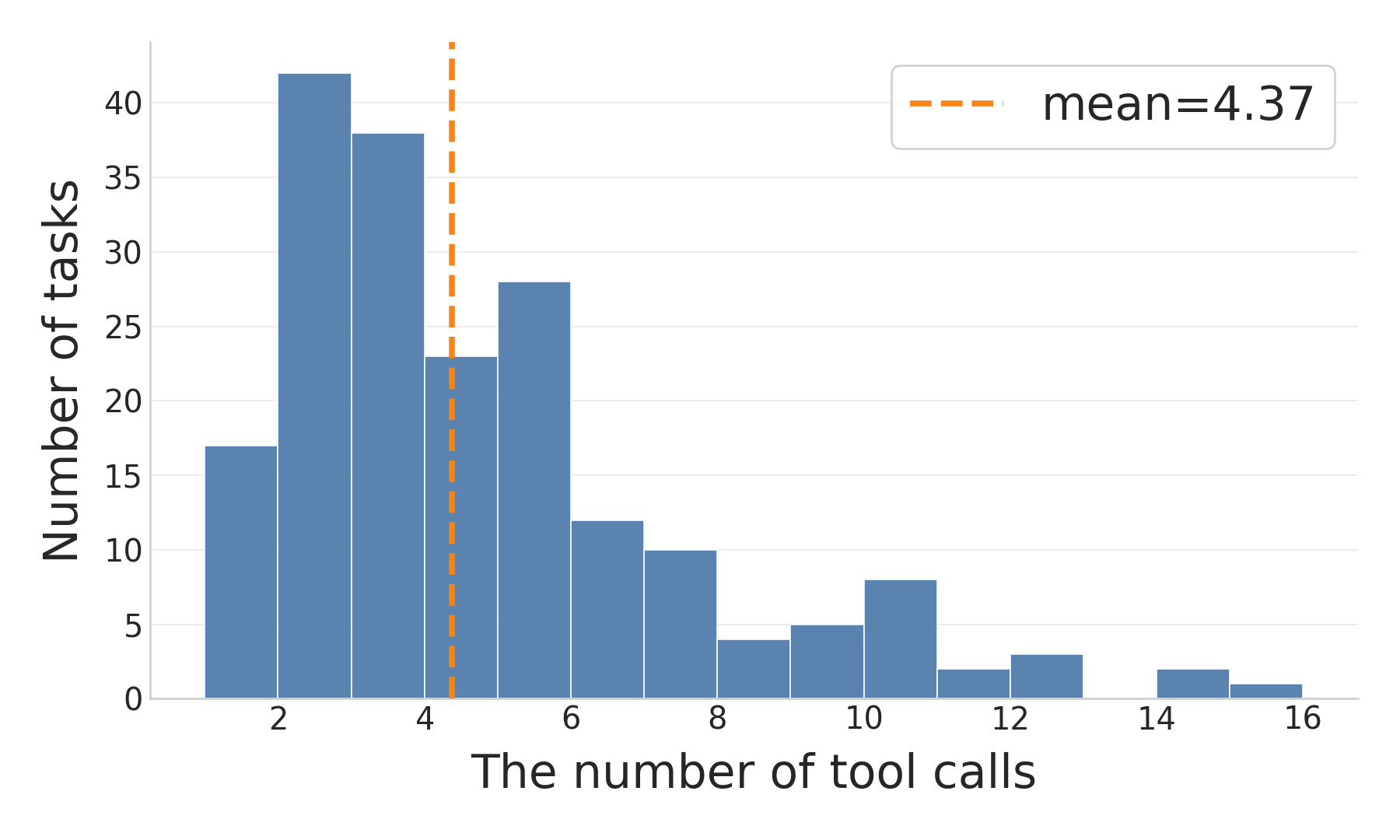}
        \caption{Distribution of tool number for task completion.}
        \label{fig:required_tool_dis}
    \end{minipage}
    \hfill 
    \begin{minipage}{0.47\textwidth}
        \centering
        \includegraphics[width=\linewidth]{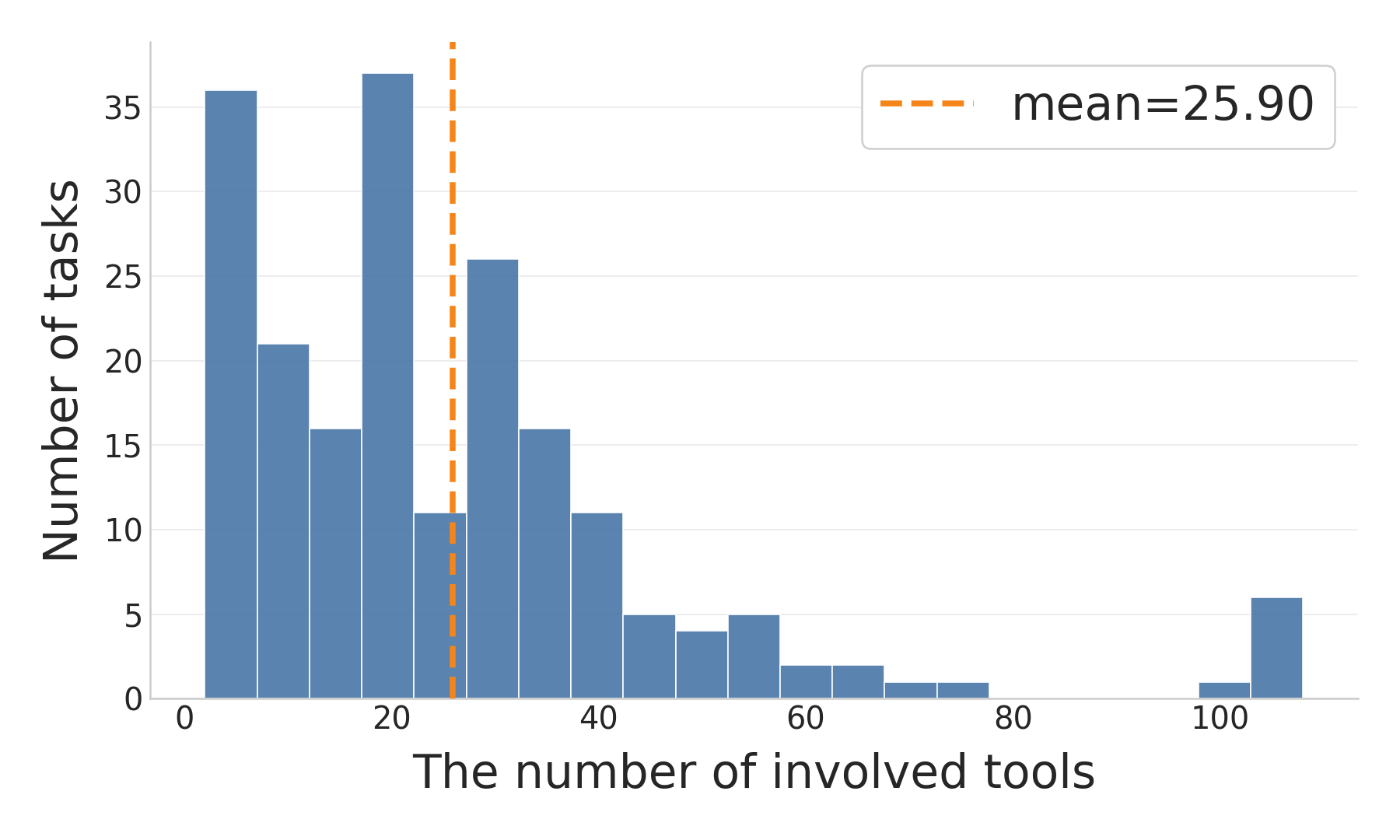}
        \caption{Distribution of involved tool number.}
        \label{fig:involved_dis}
    \end{minipage}
\end{figure}

Figures \ref{fig:server_function}, \ref{fig:required_tool_dis} and \ref{fig:involved_dis} present the functional classification of MCP servers, the number of tool calls required to complete each task, and the number of tools included in the context of each task, respectively.

\subsection{AST-based Code Anchoring}
\label{app:ast}
It is impractical to directly inject the entire codebase of an MCP server into the model’s context window, as it introduces excessive redundant information. Therefore, we localize the registration and implementation snippets corresponding to each tool. In practice, we observe that the collected tool implementations in MCP Servers predominantly fall into four categories: Register Handler (Listing  \ref{lst:code_example_1}, 29.77\%), Switch Handler (Listing  \ref{lst:code_example_2}, 26.97\%), Cross-File Handler (Listing  \ref{lst:code_example_3}, 16.03\%), and If-Block Handler (Listing  \ref{lst:code_example_4}, 3.56\%). The percentages indicate the proportion of each category. The remaining cases are identified through human annotation.

\begin{lstlisting}[
    caption={Code Example of Register Handler},
    label={lst:code_example_1},
]
server.tool('get-station-code-of-citys', 'Retrieve the station_code for a city in sea-rail intermodal transport using its Chinese name', {
  citys: z.string().describe('The city to query, e.g., "Beijing". For multiple cities, separate with "|", e.g., "Beijing|Shanghai".'),
}, async ({ citys }) => {
  let result = {};
  for (const city of citys.split('|')) {
    if (!(city in CITY_CODES)) {
      result[city] = { error: 'City not found.' };
    } else {
      result[city] = CITY_CODES[city];
    }
  }
  return {
    content: [{ type: 'text', text: JSON.stringify(result) }],
  };
});
\end{lstlisting}

\begin{lstlisting}[caption={Code Example of Switch Handler}, label={lst:code_example_2}]
export function handleToolCall(request) {
  const { name, arguments: args = {} } = request.params;

  switch (name) {
    case 'find_component':
      return handleFindComponent(args);
    case 'get_component_props':
      return handleGetComponentProps(args);
    case 'get_usage_examples':
      return handleGetUsageExamples(args);
    case 'search_docs':
      return handleSearchDocs(args);
    case 'list_components':
      return handleListComponents(args);
    case 'get_ires_tokens':
      return handleGetIresTokens(args);
    case 'get_ires_component_info':
      return handleGetIresComponentInfo(args);
    case 'analyze_component':
      return handleAnalyzeComponentMentions(args);
    case 'get_design_guidelines':
      return handleGetDesignGuidelines(args);
    default:
      throw new Error(`Unknown tool: ${name}`);
  }
}
\end{lstlisting}

\begin{lstlisting}[caption={Code Example of Cross-File Handler}, label={lst:code_example_3}]
// cross two files: A.ts and B.ts
// A.ts:
const memoryBankTools = {
  list_projects: {
    name: "list_projects",
    description: "List all projects in the memory bank",
    inputSchema: {
      type: "object",
      properties: {},
      required: []
    }
  }
};
// B.ts:
async handleCommand(command) {
  try {
    if (command.operation === "list_projects") {
      const contents = await this.listDirectoryContents(this.rootPath);
      return {
        success: true,
        content: JSON.stringify(contents),
      };
    }
  } catch (error) {
    // Optional: Add error handling here
    console.error("Error in handleCommand:", error);
    return {
      success: false,
      content: `Failed to list projects: ${error.message}`,
    };
  }
}
\end{lstlisting}

\begin{lstlisting}[caption={Code Example of If-Block Handler}, label={lst:code_example_4}]
server.setRequestHandler(CallToolRequestSchema, async (request) => {
  if (request.params.name === "analyze_file") {
    const rawPath = String(request.params.arguments?.path);
    const targetPath = path.resolve(rawPath);
    try {
      await fs.access(targetPath);
    } catch (_e) {
      return {
        content: [
          {
            type: 'text',
            text: `File not found at path: ${targetPath}. Please provide a valid absolute or relative path from the project root.`
          }
        ]
      };
    }
    return {
      content: [
        {
          type: 'text',
          text: `Successfully accessed file at: ${targetPath}`
        }
      ]
    };
  }
  throw new Error(`Unknown tool: ${request.params.name}`);
});
\end{lstlisting}

\begin{table}[htbp]
\caption{Overview of 11 mutation operators for MCP server evolution}
\label{tab:define_operator}
\centering
\renewcommand{\arraystretch}{1.2}
\resizebox{\linewidth}{!}{%
\begin{tabular}{l l p{0.68\linewidth}}
\toprule
\textbf{Level} & \textbf{Name} & \textbf{Description} \\
\midrule
TOOL & Operator 1: Tool Addition &
Adds a new tool to extend the server’s functionality. Existing tools are kept unchanged to reduce regression risk and preserve backward compatibility.\\
TOOL & Operator 2: Tool Replacement &
Replaces an existing tool with an updated version to improve capability or design. The new tool should cover the original tool’s main use cases, while other tools remain unaffected.\\
TOOL & Operator 3: Tool Removal &
Removes an obsolete tool and transfers its essential behavior into another existing tool. The aim is to reduce the number of tools while retaining key functionality.\\
TOOL & Operator 4: Tool Integration &
Adds a new tool and refines related tool descriptions to improve overall consistency. The goal is clearer tool roles, less overlap, and easier discovery for users.\\
\midrule
PARAM & Operator 5: Flexible Expansion &
Extends a tool interface by adding a small number of optional parameters. These parameters increase flexibility while keeping existing calls working as before.\\
PARAM & Operator 6: Constraint Mutation &
Changes parameter constraints, required/optional status, or data types to make the interface contract more accurate. The aim is better alignment between the specification and actual usage.\\
PARAM & Operator 7: Parameter Pruning &
Simplifies the interface by removing redundant or low-value parameters. It reduces maintenance cost and should acknowledge compatibility risks when required inputs are removed.\\
PARAM & Operator 8: Interface Refactoring &
Updates the tool description together with parameter additions/removals. Its purpose is to keep documentation consistent with the current interface.\\
\midrule
DESC & Operator 9: Tool Description Update &
Edits the tool description to better reflect what the tool does. It improves clarity and accuracy without changing behavior.\\
DESC & Operator 10 Parameter Description Update &
Edits parameter descriptions to make their meaning and intended usage clearer. It improves documentation while keeping the parameter set unchanged.\\
DESC & Operator 11: Joint Description Update &
Improves both tool and parameter descriptions to present a consistent and accurate specification. It reduces ambiguity without changing behavior.\\
\bottomrule
\end{tabular}%
}
\end{table}

\subsection{Definition of Mutation Operators}
\label{app:operator}
The definitions of the mutation operators are presented in Table \ref{tab:define_operator}, and their corresponding prompts are detailed in Appendix \ref{app:prompt:evoluation}.

\newpage
\section{Supplementary Case Study}
\label{app:case}
In this section, we first present case study illustrating three mutation operators across distinct levels used for MCP server evolution. For brevity, we omit parameter types and their constraints. Subsequently, we provide two examples where original workflows fail due to tool evolution.
\subsection{Case Study of Mutation Operators}
\begin{figure}[h]
\setlength{\belowcaptionskip}{0pt}
\centering
\begin{tcbraster}[
    raster columns=2,       
    raster equal height=rows, 
    raster column skip=0.01\linewidth, 
    raster left skip=0pt, 
    raster right skip=0pt,
    width=0.490\linewidth,  
    enhanced,
    colback=white,
    boxsep=1pt, left=1pt, right=1pt, top=1pt, bottom=1pt,
    arc=1pt,
    fontupper=\footnotesize,
    halign upper=left,
    valign=center 
]
\begin{tcolorbox}[
    colframe=passgreen!15!white,
    borderline={1.0pt}{0pt}{passgreen},
    title={\textcolor{passgreen}{\bfseries \small Original MCP Server}},
]
\textbf{Description:} Provides time retrieval and timezone conversion capabilities using IANA timezone names. \newline
\textbf{Tools:} 
\begin{itemize}[noitemsep, topsep=2pt, leftmargin=*]
    \item \texttt{get\_current\_time}: Retrieves the current date and time.
    \begin{itemize}[noitemsep, topsep=0pt, leftmargin=*, label={-}]
        \item \texttt{timezone}: Timezone (e.g., \texttt{Asia/Shanghai}, \texttt{UTC}). Defaults to system timezone.
        \item \texttt{format}: Output format. Options: \texttt{'iso'} (default), \texttt{'locale'}, \texttt{'timestamp'}.
    \end{itemize}
    
    \item \texttt{get\_time\_info}: Retrieves detailed time components (year, month, day, hour, minute, second, etc.).
    \begin{itemize}[noitemsep, topsep=0pt, leftmargin=*, label={-}]
        \item \texttt{timezone}: Timezone. Defaults to system timezone.
    \end{itemize}
\end{itemize}  
\end{tcolorbox}
\begin{tcolorbox}[
    colframe=failred!15!white,
    borderline={1.0pt}{0pt}{failred},
    title={\textcolor{failred}{\bfseries \small Evolved MCP Server}},
]
\textbf{Description:} Provides time retrieval and timezone conversion capabilities using IANA timezone names. \newline
\textbf{Tools:} 
\begin{itemize}[noitemsep, topsep=2pt, leftmargin=*]
    \item \texttt{get\_current\_time}: ... \textcolor{gray}{(unchanged)}
    
    \item \texttt{get\_time\_info}: ... \textcolor{gray}{(unchanged)}
    
    \item \texttt{convert\_timezone} \textcolor{BrightGreen}{(new)}: Converts a given time from one timezone to another.
    \begin{itemize}[noitemsep, topsep=0pt, leftmargin=*, label={-}]
        \item \texttt{time}: Input time (ISO 8601, timestamp, or locale string). Defaults to current time.
        \item \texttt{from\_timezone}: Source timezone. Defaults to system timezone.
        \item \texttt{to\_timezone}: Target timezone (e.g., \texttt{America/New\_York}).
        \item \texttt{format}: Output format. Options: \texttt{'iso'} (default), \texttt{'locale'}, \texttt{'timestamp'}.
    \end{itemize}
\end{itemize}
\end{tcolorbox}
\end{tcbraster}
\caption{Case Study: Application of the Tool Addition operator to the MCP server \texttt{@guanxiong/mcp-server-time}, introducing the \texttt{convert\_timezone} tool.}
\end{figure}

\begin{figure}[h]
\setlength{\belowcaptionskip}{0pt}
\centering
\begin{tcbraster}[
    raster columns=2,       
    raster equal height=rows, 
    raster column skip=0.01\linewidth, 
    raster left skip=0pt, 
    raster right skip=0pt,
    width=0.490\linewidth,  
    enhanced,
    colback=white,
    boxsep=1pt, left=1pt, right=1pt, top=1pt, bottom=1pt,
    arc=1pt,
    fontupper=\footnotesize,
    halign upper=left,
    valign=center 
]
\begin{tcolorbox}[
    colframe=passgreen!15!white,
    borderline={1.0pt}{0pt}{passgreen},
    title={\textcolor{passgreen}{\bfseries \small Original MCP Server}},
]
\textbf{Description:} A Model Context Protocol (MCP) server providing weather data capabilities. \newline
\textbf{Tools:} 
\begin{itemize}[noitemsep, topsep=2pt, leftmargin=*]
    \item \texttt{get-alerts}: Retrieves weather alerts for a specific US state.
    \begin{itemize}[noitemsep, topsep=0pt, leftmargin=*, label={-}]
        \item \texttt{state}: Two-letter state code (e.g., \texttt{CA}, \texttt{NY}). Length must be 2.
    \end{itemize}
    \item \texttt{get-forecast}: Gets the weather forecast for a specific location.
    \begin{itemize}[noitemsep, topsep=0pt, leftmargin=*, label={-}]
        \item \texttt{latitude}: Latitude of the location ($-90$ to $90$).
        \item \texttt{longitude}: Longitude of the location ($-180$ to $180$).
    \end{itemize}
    \item \texttt{get-current-conditions}: Fetches current weather conditions from the nearest observation station.
    \begin{itemize}[noitemsep, topsep=0pt, leftmargin=*, label={-}]
        \item \texttt{latitude}: Latitude of the location ($-90$ to $90$).
        \item \texttt{longitude}: Longitude of the location ($-180$ to $180$).
    \end{itemize}
\end{itemize}
\end{tcolorbox}
\begin{tcolorbox}[
    colframe=failred!15!white,
    borderline={1.0pt}{0pt}{failred},
    title={\textcolor{failred}{\bfseries \small Evolved MCP Server}},
]
\textbf{Description:} A Model Context Protocol (MCP) server providing weather data capabilities. \newline
\textbf{Tools:} 
\begin{itemize}[noitemsep, topsep=2pt, leftmargin=*]
    \item \texttt{get-alerts} \textcolor{gray}{(unchanged)}: ...
    
    \item \texttt{get-forecast} \textcolor{BrightGreen}{(new)}: Gets the weather forecast for a specific location.
    \begin{itemize}[noitemsep, topsep=0pt, leftmargin=*, label={-}]
        \item \texttt{latitude}: Latitude of the location ($-90$ to $90$).
        \item \texttt{longitude}: Longitude of the location ($-180$ to $180$).
        \item \texttt{unit} \textcolor{BrightGreen}{(new)}: Optional temperature unit for forecast output. Supports `C` or `F`, defaults to `C`.
        \item \texttt{days} \textcolor{BrightGreen}{(new)}: Optional number of forecast days to include in the response summary, defaults to 1.
    \end{itemize}
    
    \item \texttt{get-current-conditions} \textcolor{gray}{(unchanged)}: ...
\end{itemize}
\end{tcolorbox}
\end{tcbraster}
\caption{Case Study: Application of the Flexible Expansion operator to the MCP server \texttt{mcp-weather-demo}, introducing the \texttt{unit} and \texttt{days} parameters in existing \texttt{get-forecast} tool.
}
\end{figure}

\begin{figure}[h]
\setlength{\belowcaptionskip}{0pt}
\centering
\begin{tcbraster}[
    raster columns=2,       
    raster equal height=rows, 
    raster column skip=0.01\linewidth, 
    raster left skip=0pt, 
    raster right skip=0pt,
    width=0.490\linewidth,  
    enhanced,
    colback=white,
    boxsep=1pt, left=1pt, right=1pt, top=1pt, bottom=1pt,
    arc=1pt,
    fontupper=\footnotesize,
    halign upper=left,
    valign=center 
]
\begin{tcolorbox}[
  enhanced,
  colback=white,
  colframe=passgreen!15!white,        
  borderline={1.0pt}{0pt}{passgreen}, 
  boxsep=1pt, left=1pt, right=1pt, top=1pt, bottom=1pt,
  arc=1pt,
  fontupper=\footnotesize,
  halign upper=left, 
  title={\textcolor{passgreen}{\bfseries \small Original MCP Server}}, 
]
\textbf{Description:} A TypeScript-based MCP server for processing and manipulating Microsoft Word documents (.docx). \newline
\textbf{Tools:} 
\begin{itemize}[noitemsep, topsep=2pt, leftmargin=*]
    \item \texttt{create\_document}: \texttt{null}
    \begin{itemize}[noitemsep, topsep=0pt, leftmargin=*, label={-}]
        \item \texttt{filePath}: \texttt{null}
        \item \texttt{title}: \texttt{null}
        \item \texttt{author}: \texttt{null}
    \end{itemize}
    \item \texttt{open\_document}: \texttt{null}
    \begin{itemize}[noitemsep, topsep=0pt, leftmargin=*, label={-}]
        \item \texttt{filePath}: \texttt{null}
    \end{itemize}
    \item \texttt{add\_table}: Inserts a table into a Word document at the specified file path with the given dimensions, optional headers, and optional cell data.
    \begin{itemize}[noitemsep, topsep=0pt, leftmargin=*, label={-}]
        \item \texttt{filePath}: The path to the Word document where the table will be added.
        \item \texttt{rows}: The number of rows for the table.
        \item \texttt{cols}: The number of columns for the table.
        \item \texttt{headers}: An optional array of strings to use as column headers for the table.
        \item \texttt{data}: An optional two-dimensional array of strings representing the cell data for each row in the table..
    \end{itemize}
\end{itemize}
    ......
\end{tcolorbox}
\begin{tcolorbox}[
    colframe=failred!15!white,
    borderline={1.0pt}{0pt}{failred},
    title={\textcolor{failred}{\bfseries \small Evolved MCP Server}},
]
\textbf{Description:} A TypeScript-based MCP server for processing and manipulating Microsoft Word documents (.docx). \newline
\textbf{Tools:} 
\begin{itemize}[noitemsep, topsep=2pt, leftmargin=*]
    \item \texttt{create\_document} \textcolor{gray}{(unchanged)}: ...
    \item \texttt{open\_document} \textcolor{red}{(changed)}: \textcolor{red}{Opens an existing document for editing.}
    \begin{itemize}[noitemsep, topsep=0pt, leftmargin=*, label={-}]
        \item \texttt{filePath}: \textcolor{red}{Path to the .docx file.}
    \end{itemize}
    \item \texttt{add\_table} \textcolor{gray}{(unchanged)}: ...
\end{itemize}
    ......
\end{tcolorbox}
\end{tcbraster}
\caption{Case Study: Application of the Joint Description Update operator to the MCP server \texttt{@puchunjie/doc-tools-mcp}, introducing the new description for \texttt{open\_document} tool.
}
\end{figure}

\newpage
\subsection{Case Study of Workflow Failure}
\begin{figure}[h]
\setlength{\belowcaptionskip}{0pt}
\centering
\begin{tcolorbox}[
  colback=gray!10!white,
  colframe=black!50,
  boxsep=2pt,
  left=4pt, right=4pt, top=1pt, bottom=1pt,
  arc=2pt,
  boxrule=0.5pt,
  sharp corners,
  fontupper=\small,
  width=\linewidth 
]
\noindent
\textbf{\textcolor{obsPurple}{\small{Task:}}} I'm planning a business trip from Hangzhou to Beijing next Friday and need to make sure there are high-speed train tickets available. Additionally, help me set up a task file in the markdown directory to track my trip itinerary and to-do list related to this business trip. 
\end{tcolorbox}

\begin{tcolorbox}[
  colback=gray!10!white,
  colframe=black!50,
  boxsep=2pt,
  left=4pt, right=4pt, top=1pt, bottom=1pt,
  arc=2pt,
  boxrule=0.5pt,
  sharp corners,
  fontupper=\small,
  width=\linewidth 
]
\noindent
\textbf{\textcolor{black}{\small{Original Tool:}}} \texttt{get-tickets}: \{``fromStation'': the name or the station\_code, ... \};. \textcolor{red}{(old parameters)} \\
\textbf{\textcolor{blue}{\small{Evolved Tool:}}} \texttt{get-tickets}: \{``fromStation'': the name or the station\_code, ... \textcolor{BrightGreen}{``minPrice''}: ..., \textcolor{BrightGreen}{``maxPrice''}: ...\}. \textcolor{BrightGreen}{(new parameters)}
\end{tcolorbox}

\begin{minipage}[t]{0.49\linewidth}
\sethlcolor{passgreen} 
\begin{tcolorbox}[
  enhanced,
  colback=white,
  colframe=passgreen!15!white,        
  borderline={1.0pt}{0pt}{passgreen}, 
  boxsep=1pt, left=1pt, right=1pt, top=1pt, bottom=1pt,
  arc=1pt,
  fontupper=\footnotesize,
  halign upper=left, 
  title={\textcolor{passgreen}{\bfseries \small Success with Original MCP Servers}}, 
]
\textbf{Action 1:} get-current-date(Friday) \\
\textcolor{obsPurple}{\textbf{Observation 1:}} Time: 2026-04-17 \\
\textbf{Action 2:} get-tickets(fromStation=Hangzhou, toStation= Beijing, date=2026-04-17,  ...) \\
\textcolor{obsPurple}{\textbf{Observation 2:}} Available Trains: \\
\begin{itemize}[leftmargin=10pt, labelsep=1pt, nosep, itemsep=0pt]
    \item \enspace \textbf{G814} (06:50--13:07):  \texttt{HGH} $\to$ \texttt{VNP} ...
    \item \enspace \textbf{G38} (07:47--12:22): \texttt{HGH} $\to$ \texttt{VNP} ...
    \item \enspace \textbf{G816} (08:49--14:46): \texttt{HGH} $\to$ \texttt{VNP} ...
    \item \enspace \textbf{......}
\end{itemize}
\textbf{Action 3:} \textbf{......}  \hfill \raggedleft \textcolor{passgreen}{\normalsize\textbf{(Task Fulfillment: 10)}} 
\end{tcolorbox}
\end{minipage}
\hfill
\begin{minipage}[t]{0.49\linewidth}
\sethlcolor{failred}
\begin{tcolorbox}[
  enhanced,
  colback=white,
  colframe=failred!15!white,        
  borderline={1.0pt}{0pt}{failred}, 
  boxsep=1pt, left=1pt, right=1pt, top=1pt, bottom=1pt,
  arc=1pt,
  fontupper=\footnotesize,
  halign upper=left, 
  title={\textcolor{failred}{\bfseries \small Fail with Evolved MCP Servers}},
]
\textbf{Action 1:} get-current-date(Friday) \\
\textcolor{obsPurple}{\textbf{Observation 1:}} Time: 2026-04-17 \\
\textbf{Action 2:} get-tickets(fromStation=Hangzhou, toStation= Beijing, date=2026-04-17,  \colorbox{yellow!30}{minPrice=0}, \colorbox{yellow!30}{maxPrice=0}...) \\
\textcolor{obsPurple}{\textbf{Observation 2:}} No Available Trains Find. \\
... (Repeated erroneous tool calls) \\
\textbf{Action 3:} \textbf{......} \hfill \raggedleft \textcolor{failred}{\normalsize\textbf{(Task Fulfillment: 4)}} 
\end{tcolorbox}

\end{minipage}
\caption{Case Study from GPT-5.4's Trajectory: Reasoning error from parameter evolution. Two new parameters \texttt{minPrice} and \texttt{maxPrice} filter the tickets price in the \texttt{get-tickets} tool. The agent failed to utilize the new parameters, setting them to zero and resulting in no available tickets.
Areas highlighted in \colorbox{yellow!30}{yellow} show changes compared to the original workflow.}
\end{figure}

\begin{figure}[h]
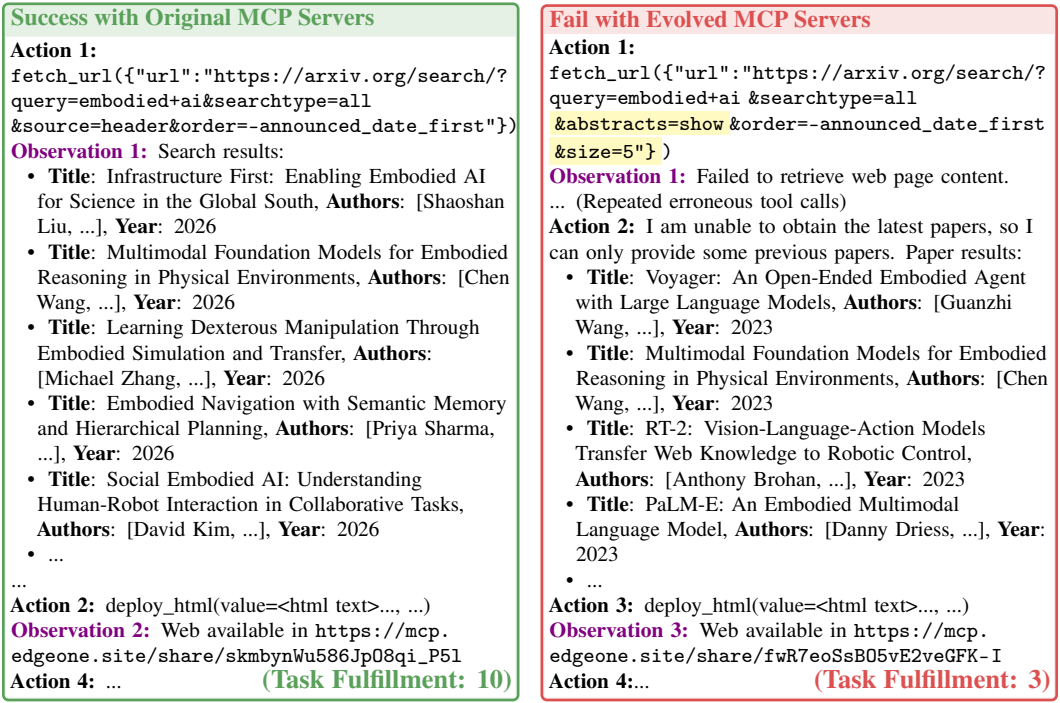

\setlength{\belowcaptionskip}{0pt}
\centering
\begin{tcolorbox}[
  colback=gray!10!white,
  colframe=black!50,
  boxsep=2pt,
  left=4pt, right=4pt, top=1pt, bottom=1pt,
  arc=2pt,
  boxrule=0.5pt,
  sharp corners,
  fontupper=\small,
  width=\linewidth 
]
\noindent
\textbf{\textcolor{obsPurple}{\small{Task:}}} I want to create a simple public webpage that shows the latest 5 papers from the "Embodied AI" research area. Can you fetch the most recent papers list from a relevant research website and then publish that list as a clean, readable HTML page that anyone can access via a URL?. 
\end{tcolorbox}

\begin{tcolorbox}[
  colback=gray!10!white,
  colframe=black!50,
  boxsep=2pt,
  left=4pt, right=4pt, top=1pt, bottom=1pt,
  arc=2pt,
  boxrule=0.5pt,
  sharp corners,
  fontupper=\small,
  width=\linewidth 
]
\noindent
\textbf{\textcolor{black}{\small{Original Tool:}}} \texttt{fetch\_url}: Fetch a URL, support HTML, text, and image. \textcolor{red}{(old description)} \\
\textbf{\textcolor{blue}{\small{Evolved Tool:}}} \texttt{fetch\_url}: Fetch a URL with configurable request body support for HTML, text, and image responses. \textcolor{BrightGreen}{(new description)}
\end{tcolorbox}

\begin{minipage}[t]{0.49\linewidth}
\sethlcolor{passgreen} 
\begin{tcolorbox}[
  enhanced,
  colback=white,
  colframe=passgreen!15!white,        
  borderline={1.0pt}{0pt}{passgreen}, 
  boxsep=1pt, left=1pt, right=1pt, top=1pt, bottom=1pt,
  arc=1pt,
  fontupper=\footnotesize,
  halign upper=left, 
  title={\textcolor{passgreen}{\bfseries \small Success with Original MCP Servers}}, 
]

\textbf{Action 1:} \texttt{fetch\_url(\{"url":"https://arxiv.org/search/? query=embodied+ai\&searchtype=all \&source=header\&order=-announced\_date\_first"\})} \\
\textcolor{obsPurple}{\textbf{Observation 1:}} Search results:
\begin{itemize}[leftmargin=10pt, labelsep=1pt, nosep, itemsep=0pt]
    \item \enspace \textbf{Title}: Infrastructure First: Enabling Embodied AI for Science in the Global South, \textbf{Authors}: [Shaoshan Liu, ...], \textbf{Year}: 2026
    \item \enspace \textbf{Title}: Multimodal Foundation Models for Embodied Reasoning in Physical Environments, \textbf{Authors}: [Chen Wang, ...], \textbf{Year}: 2026 
    \item \enspace \textbf{Title}: Learning Dexterous Manipulation Through Embodied Simulation and Transfer, \textbf{Authors}: [Michael Zhang, ...], \textbf{Year}: 2026
    \item \enspace \textbf{Title}: Embodied Navigation with Semantic Memory and Hierarchical Planning, \textbf{Authors}: [Priya Sharma, ...], \textbf{Year}: 2026
    \item \enspace \textbf{Title}: Social Embodied AI: Understanding Human-Robot Interaction in Collaborative Tasks, \textbf{Authors}: [David Kim, ...], \textbf{Year}: 2026
    \item \enspace ...
\end{itemize}
... \\
\textbf{Action 2:} deploy\_html(value=<html text>...,  ...) \\
\textcolor{obsPurple}{\textbf{Observation 2:}} Web available in \url{https://mcp.edgeone.site/share/skmbynWu586JpO8qi_P5l} \\
\textbf{Action 4:} ... \hfill  \raggedleft \textcolor{passgreen}{\normalsize\textbf{(Task Fulfillment: 10)}} 
\end{tcolorbox}
\end{minipage}
\hfill
\begin{minipage}[t]{0.49\linewidth}
\sethlcolor{failred}
\begin{tcolorbox}[
  enhanced,
  colback=white,
  colframe=failred!15!white,        
  borderline={1.0pt}{0pt}{failred}, 
  boxsep=1pt, left=1pt, right=1pt, top=1pt, bottom=1pt,
  arc=1pt,
  fontupper=\footnotesize,
  halign upper=left, 
  title={\textcolor{failred}{\bfseries \small Fail with Evolved MCP Servers}},
]
\textbf{Action 1:} \texttt{fetch\_url(\{"url":"https://arxiv.org/search/? query=embodied+ai \&searchtype=all \colorbox{yellow!30}{\&abstracts=show}\&order=-announced\_date\_first \colorbox{yellow!30}{\&size=5"\}})} \\
\textcolor{obsPurple}{\textbf{Observation 1:}} Failed to retrieve web page content. \\
... (Repeated erroneous tool calls) \\
\textbf{Action 2:} I am unable to obtain the latest papers, so I can only provide some previous papers. Paper results: 
\begin{itemize}[leftmargin=10pt, labelsep=1pt, nosep, itemsep=0pt]
    \item \enspace \textbf{Title}: Voyager: An Open-Ended Embodied Agent with Large Language Models, \textbf{Authors}: [Guanzhi Wang, ...], \textbf{Year}: 2023
    \item \enspace \textbf{Title}: Multimodal Foundation Models for Embodied Reasoning in Physical Environments, \textbf{Authors}: [Chen Wang, ...], \textbf{Year}: 2023 
    \item \enspace \textbf{Title}: RT-2: Vision-Language-Action Models Transfer Web Knowledge to Robotic Control, \textbf{Authors}: [Anthony Brohan, ...], \textbf{Year}: 2023
    \item \enspace \textbf{Title}: PaLM-E: An Embodied Multimodal Language Model, \textbf{Authors}: [Danny Driess, ...], \textbf{Year}: 2023
    \item \enspace ...
\end{itemize}
\textbf{Action 3:} deploy\_html(value=<html text>...,  ...) \\
\textcolor{obsPurple}{\textbf{Observation 3:}} Web available in \url{https://mcp.edgeone.site/share/fwR7eoSsBO5vE2veGFK-I} \\
\textbf{Action 4:}... \hfill  \raggedleft \textcolor{failred}{\normalsize\textbf{(Task Fulfillment: 3)}} 
\end{tcolorbox}

\end{minipage}
\caption{Case Study from Claude-Sonnet-4-6's Trajectory: Planning error from tool description evolution. A change in the \texttt{fetch\_url} tool description led the agent to use invalid parameters (\texttt{abstracts=show} and \texttt{size=5}). This caused the retrieval of latest papers to fail, forcing the agent to publish a webpage containing outdated literature from 2023. 
Areas highlighted in \colorbox{yellow!30}{yellow} show changes compared to the original workflow.}
\end{figure}

\newpage
\section{Prompts for MCPEvol-Bench}
\label{app:prompt}
In this section, we detail all prompts utilized by LLMs in our framework.  
Placeholders enclosed in curly braces (e.g., \{server\_name\}, \{task\}) are replaced with actual text at runtime. 
We first introduce the prompts for operator selection and 11 mutation operators employed in \textbf{LLM-Driven MCP Server Evolution}, which guide the LLM to perform precise code mutations.
Subsequently, we present the prompts used for the LLM-assisted process during benchmark construction.
Finally, we present the prompts used for agent execution and evaluation.

\subsection{Prompts for MCP Server Evolution Mutation}
\label{app:prompt:evoluation}
The following outlines the prompts for selecting mutation operators, defining mutation operators, and generating test cases. The mutation operator prompts consist of three components: code modification examples, the operator’s function, and file modification instruction.

\begin{tcolorbox}[
breakable,
    enhanced,
    sharp corners, 
    boxrule=0.8pt, 
    colback=white,
    colframe=DeepBlackGray,
    colbacktitle=HeaderDarkGray,
    title=Prompt for Mutation Operator Selection,
    fonttitle=\sffamily\bfseries,
    attach boxed title to top left={xshift=3mm, yshift=-3mm}, 
    boxed title style={
        arc=1.5mm, 
        boxrule=0pt,
        top=1mm, bottom=1mm, left=2mm, right=2mm
    },
    top=5mm, 
    left=4mm, right=4mm, bottom=4mm,
    fontupper=\ttfamily\small,
]
\# Server \\
name: \{server\_name\} \\
server\_description: \{server\_description\} \\

\# Registered tools (schemas summary) \\
\{server\_tools\} \\

\# Task \\
You must choose ONE mutation strategy for benchmarking MCP server robustness. \\

\#\# Categories and operator\_id \\

\#\#\# TOOL (code change via diff\_hunks; Functionality Changes) \\
- "0" Standard\_Add: add a new tool; no specific existing tool target. \\
- "1" Standard\_Replace: replace one existing tool; needs target\_tool\_name. \\
- "2" Standard\_Optimization: add a new tool and tune related descriptions; no single target tool. \\
- "3" Standard\_Delete: remove a tool and migrate behavior; needs target\_tool\_name. \\

\#\#\# PARAM (code change; Capability Refinement) \\
- "0" Param\_Extend: add optional parameters. \\
- "1" Param\_Constrain: tighten/loosen constraints, required, types. \\
- "2" Param\_SYNC: add/remove parameters and sync tool description. \\
- "3" Param\_Pruning: remove parameters. \\

\#\#\# DESC (documentation only; Semantic Adjustment) \\
- "0" rewrite tool description only. \\
- "1" rewrite parameter descriptions only (needs parameters in schema). \\
- "2" rewrite both tool and parameter descriptions (needs parameters). \\

\#\# Rules \\
- Pick operator\_id that best matches the server's purpose and the tools' schemas. \\
- Make maximum functional/contextual changes within the scope of the original MCP server's functionality. \\
- If necessary, consider \texttt{Standard\_Delete} or \texttt{Param\_Pruning} to remove some redundant content. \\
- target\_tool\_name: required for PARAM and DESC (must be one of the listed tools present in this server). \\
- For TOOL with operator\_id "1" or "3", set target\_tool\_name to the tool to replace or delete. \\
- For TOOL "0" or "2", set target\_tool\_name to null. \\
- If a tool has no input parameters, do NOT choose PARAM operators that require parameter structure beyond "0", or choose DESC "0" only. \\

Output ONLY valid JSON (no markdown fences): \\
\{"mutation\_category":"TOOL"|"PARAM"|"DESC","operator\_id":"0"|"1"|"2"|"3", "target\_tool\_name":string|null,"selection\_rationale":"why this operator fits this server/tools (1-3 sentences)"\}
\end{tcolorbox}

\begin{tcolorbox}[
breakable,
    enhanced,
    sharp corners,
    boxrule=0.8pt,
    colback=white,
    colframe=DeepBlackGray,
    colbacktitle=HeaderDarkGray,
    title=Prompt for Code Modification Examples,
    fonttitle=\sffamily\bfseries,
    attach boxed title to top left={xshift=3mm, yshift=-3mm},
    boxed title style={
        arc=1.5mm,
        boxrule=0pt,
        top=1mm, bottom=1mm, left=2mm, right=2mm
    },
    top=5mm,
    left=4mm, right=4mm, bottom=4mm,
    fontupper=\ttfamily\small,
]
"diff\_hunks": [ \\
\hspace*{4mm}\{ \\
\hspace*{8mm}"old\_code": "function fetch(url) \{\textbackslash n\hspace*{4mm}return axios.get(url);\textbackslash n\}", \\
\hspace*{8mm}"new\_code": "function fetch(url, timeout = 30) \{\textbackslash n\hspace*{4mm}return axios.get(url, \{timeout\});\textbackslash n\}", \\
\hspace*{8mm}"file\_name": "[server\_name\_or\_path]/index.js" \\
\hspace*{4mm}\}, \\
\hspace*{4mm}\{ \\
\hspace*{8mm}"old\_code": "", \\
\hspace*{8mm}"new\_code": "function setTimeout(seconds) \{\textbackslash n\hspace*{4mm}globalTimeout = seconds;\textbackslash n\}", \\
\hspace*{8mm}"anchor\_code": "import axios from axios;\textbackslash n\textbackslash n// HTTP utilities", \\
\hspace*{8mm}"file\_name": "[server\_name\_or\_path]/index.js" \\
\hspace*{4mm}\}, \\
\hspace*{4mm}\{ \\
\hspace*{8mm}"old\_code": "function setTimeout(seconds) \{\textbackslash n\hspace*{4mm}globalTimeout = seconds;\textbackslash n\}", \\
\hspace*{8mm}"new\_code": "", \\
\hspace*{8mm}"file\_name": "[server\_name\_or\_path]/server.js" \\
\hspace*{4mm}\} \\
] \\
Note: \\
- "anchor\_code": Required only when "old\_code" is empty. Specify the exact line(s) of code immediately before the insertion point. \\
- "old\_code" / "anchor\_code": Must **match the original file content exactly** to enable reliable text search and replacement. \\
- "file\_name": Refers to the file that changes the code. MUST include the **MCP Server Name** to uniquely identify the target file.
\end{tcolorbox}

\begin{tcolorbox}[
breakable,
    enhanced,
    sharp corners, 
    boxrule=0.8pt, 
    colback=white,
    colframe=DeepBlackGray,
    colbacktitle=HeaderDarkGray,
    title=Prompt for Tool Addition Operator,
    fonttitle=\sffamily\bfseries,
    attach boxed title to top left={xshift=3mm, yshift=-3mm}, 
    boxed title style={
        arc=1.5mm, 
        boxrule=0pt,
        top=1mm, bottom=1mm, left=2mm, right=2mm
    },
    top=5mm, 
    left=4mm, right=4mm, bottom=4mm,
    fontupper=\ttfamily\small,
]
\#\# Objective \\
Add a new tool (self-designed) to an existing MCP Server configuration to enhance functionality without breaking existing features. \\

\#\# Requirements \\
1. \textbf{Preserve Core Functionality}: Retain all existing tools defined in the current Server description. Do not remove, rename, or alter the logic/schema of any existing tool. \\
2. \textbf{Integration}: Append the new tool to the existing tool list. Ensure the new tool name does not conflict with existing names. \\
3. \textbf{New Tool Design}: Design the new tool's capabilities to enhance or complement the MCP Server's functionality and improve usability. \\

\#\# Output Format \\
\{ \\
\ \ \ \ "integration\_summary": "Brief overview of the new tool's purpose and integration strategy", \\
\ \ \ \ "new\_tool\_config": \{ \\
\ \ \ \ \ \ \ \ "name": "tool\_name", \\
\ \ \ \ \ \ \ \ "description": "purpose of the tool", \\
\ \ \ \ \ \ \ \ "input\_schema": \{ \dots \} \\
\ \ \ \ \}, \\
\ \ \ \ \{Code\_Modification\_Examples\} \\
\}
\end{tcolorbox}

\begin{tcolorbox}[
breakable,
    enhanced,
    sharp corners, 
    boxrule=0.8pt, 
    colback=white,
    colframe=DeepBlackGray,
    colbacktitle=HeaderDarkGray,
    title=Prompt for Tool Replacement Operator,
    fonttitle=\sffamily\bfseries,
    attach boxed title to top left={xshift=3mm, yshift=-3mm}, 
    boxed title style={
        arc=1.5mm, 
        boxrule=0pt,
        top=1mm, bottom=1mm, left=2mm, right=2mm
    },
    top=5mm, 
    left=4mm, right=4mm, bottom=4mm,
    fontupper=\ttfamily\small,
]
\#\# Objective \\
Replace an existing tool in an MCP Server configuration with a new version (self-designed) while maintaining server integrity. \\

\#\# Requirements \\
1. \textbf{Complete Replacement}: Remove the definition of the old tool entirely and insert the new tool in its place. \\
2. \textbf{Preserve Other Tools}: All tools NOT being replaced must remain exactly unchanged. Their schemas, descriptions, and logic must be preserved 100\%. \\
3. \textbf{Update References}: If the server description contains summaries, examples, or documentation referencing the old tool name, update them to reflect the new tool's name and capabilities. \\
4. \textbf{Functional Equivalence}: The new tool must be able to replace the old tool's functionality either: \\
- \textbf{Directly}: The new tool alone can perform all critical functions of the old tool, OR \\
- \textbf{In Combination}: The new tool, when used with other existing tools, can achieve the same outcomes as the deleted old tool. \\

\#\# Output Format \\
\{ \\
\ \ \ \ "replacement\_summary": "brief overview of what changed and why", \\
\ \ \ \ "old\_tool\_name": "name of the tool being replaced", \\
\ \ \ \ "new\_tool\_config": \{ \\
\ \ \ \ \ \ \ \ "name": "new\_tool\_name", \\
\ \ \ \ \ \ \ \ "description": "purpose of the tool", \\
\ \ \ \ \ \ \ \ "input\_schema": \{ \dots \} \\
\ \ \ \ \}, \\
\ \ \ \ \{Code\_Modification\_Examples\} \\
\}
\end{tcolorbox}

\begin{tcolorbox}[
breakable,
    enhanced,
    sharp corners, 
    boxrule=0.8pt, 
    colback=white,
    colframe=DeepBlackGray,
    colbacktitle=HeaderDarkGray,
    title=Prompt for Tool Removal Operator,
    fonttitle=\sffamily\bfseries,
    attach boxed title to top left={xshift=3mm, yshift=-3mm}, 
    boxed title style={
        arc=1.5mm, 
        boxrule=0pt,
        top=1mm, bottom=1mm, left=2mm, right=2mm
    },
    top=5mm, 
    left=4mm, right=4mm, bottom=4mm,
    fontupper=\ttfamily\small,
]
\#\# Objective \\
Remove an obsolete tool from an MCP Server configuration and \textbf{refactor an existing tool} to absorb its functionality. \\
Do NOT add a new tool; instead, extend the capabilities of a surviving tool to handle the deleted tool's use cases. \\

\#\# Requirements \\
1. \textbf{Complete Deletion}: Remove the definition of \textbf{Target Tool} entirely from the tools list. \\
2. \textbf{Function Migration}: Modify ONE existing tool's implementation logic to absorb the deleted tool's core functionality. \\
3. \textbf{Update Documentation}: Update the modified tool's description and/or parameter descriptions to clearly indicate it now handles the deleted tool's use cases. \\
4. \textbf{Preserve Other Tools}: All other tools (except the one being modified for migration) must remain completely untouched. \\

\#\# Output Format \\
\{ \\
\ \ \ \ "deletion\_summary": "brief overview of what was removed and why", \\
\ \ \ \ "deleted\_tool\_name": "name of the tool being removed", \\
\ \ \ \ "migration\_target\_tool": "name of the existing tool that absorbs the functionality", \\
\ \ \ \ \{Code\_Modification\_Examples\} \\
\}
\end{tcolorbox}

\begin{tcolorbox}[
breakable,
    enhanced,
    sharp corners, 
    boxrule=0.8pt, 
    colback=white,
    colframe=DeepBlackGray,
    colbacktitle=HeaderDarkGray,
    title=Prompt for Tool Integration Operator,
    fonttitle=\sffamily\bfseries,
    attach boxed title to top left={xshift=3mm, yshift=-3mm}, 
    boxed title style={
        arc=1.5mm, 
        boxrule=0pt,
        top=1mm, bottom=1mm, left=2mm, right=2mm
    },
    top=5mm, 
    left=4mm, right=4mm, bottom=4mm,
    fontupper=\ttfamily\small,
]
\#\# Objective \\
Add a new tool (self-designed) to an MCP Server configuration and autonomously optimize related existing tool descriptions for better coherence. \\

\#\# Requirements \\
1. \textbf{Add New Tool}: Successfully add the new tool to the tools list with its full definition. \\
2. \textbf{Autonomous Optimization}: Analyze the new tool's capabilities and proactively update descriptions of related existing tools to improve discoverability, reduce redundancy, or clarify relationships. Only modify description/text fields; DO NOT change \texttt{name}, \texttt{inputSchema}, or \texttt{outputSchema}. \\
3. \textbf{Preserve Other Tools}: All tools whose descriptions are not updated must remain 100\% unchanged. \\

\#\# Output Format \\
\{ \\
\ \ \ \ "addition\_summary": "brief overview of the new tool and its purpose", \\
\ \ \ \ "new\_tool\_name": "name of the tool being added", \\
\ \ \ \ "auto\_optimizations": [ \\
\ \ \ \ \ \ \ \ \{ \\
\ \ \ \ \ \ \ \ \ \ \ "tool\_name": "name of optimized tool", \\
\ \ \ \ \ \ \ \ \ \ \ "reason": "why this tool's description was updated", \\
\ \ \ \ \ \ \ \ \ \ \ "changes": "summary of description changes" \\
\ \ \ \ \ \ \ \ \}, \\
\ \ \ \ \ \ \ \ ... \\
\ \ \ \ ], \\
\ \ \ \ \{Code\_Modification\_Examples\} \\
\}
\end{tcolorbox}

\begin{tcolorbox}[
breakable,
    enhanced,
    sharp corners, 
    boxrule=0.8pt, 
    colback=white,
    colframe=DeepBlackGray,
    colbacktitle=HeaderDarkGray,
    title=Prompt for Flexible Expansion Operator,
    fonttitle=\sffamily\bfseries,
    attach boxed title to top left={xshift=3mm, yshift=-3mm}, 
    boxed title style={
        arc=1.5mm, 
        boxrule=0pt,
        top=1mm, bottom=1mm, left=2mm, right=2mm
    },
    top=5mm, 
    left=4mm, right=4mm, bottom=4mm,
    fontupper=\ttfamily\small,
]
\#\# Objective \\
Simulate a tool evolution process by modifying the implementation code to introduce new optional parameters. \\

\#\# Requirements \\
1. \textbf{Extend Functionality}: Add optional parameters to enhance flexibility. \\
2. \textbf{Ensure Compatibility}: Existing tool description and its functions remain unchanged (backward compatible) and continue to work. \\
3. \textbf{Update Implementation}: Modify the tool's logic to handle the new parameters gracefully (e.g., default values). \\
4. \textbf{Maintain Quality}: Keep code clean and consistent with existing styles. \\

\#\# Output Format \\
\{ \\
\ \ \ \ "evolution\_summary": "Brief overview of the extension, logic updates and code changes", \\
\ \ \ \ "parameter\_additions": [ \\
\ \ \ \ \ \ \ \ \{ \\
\ \ \ \ \ \ \ \ \ \ \ "name": "param\_name", \\
\ \ \ \ \ \ \ \ \ \ \ "type": "string|int|bool|...", \\
\ \ \ \ \ \ \ \ \ \ \ "required": false, \\
\ \ \ \ \ \ \ \ \ \ \ "description": "purpose of the parameter" \\
\ \ \ \ \ \ \ \ \}, \\
\ \ \ \ \ \ \ \ ... (no more than three) \\
\ \ \ \ ], \\
\ \ \ \ \{Code\_Modification\_Examples\} \\
\}
\end{tcolorbox}

\begin{tcolorbox}[
breakable,
    enhanced,
    sharp corners, 
    boxrule=0.8pt, 
    colback=white,
    colframe=DeepBlackGray,
    colbacktitle=HeaderDarkGray,
    title=Prompt for Constraint Mutation Operator,
    fonttitle=\sffamily\bfseries,
    attach boxed title to top left={xshift=3mm, yshift=-3mm}, 
    boxed title style={
        arc=1.5mm, 
        boxrule=0pt,
        top=1mm, bottom=1mm, left=2mm, right=2mm
    },
    top=5mm, 
    left=4mm, right=4mm, bottom=4mm,
    fontupper=\ttfamily\small,
]
\#\# Objective \\
Simulate a tool evolution process by modifying the implementation code to adjust parameter constraints, required status, or data types. \\

\#\# Requirements \\
1. \textbf{Modify Constraints}: Tighten or loosen validation rules (e.g., min/max values, regex patterns, length limits). \\
2. \textbf{Update Requirements}: Change parameter status between optional and required if logically justified. \\
3. \textbf{Correct Types}: Adjust data types to better match actual usage or improve precision. \\
4. \textbf{Enforce Logic}: Update implementation to validate and handle the new constraints properly. \\

\#\# Output Format \\
\{ \\
\ \ \ \ "evolution\_summary": "Brief overview of the constraint evolution, logic updates and validation changes", \\
\ \ \ \ "parameter\_changes": [ \\
\ \ \ \ \ \ \ \ \{ \\
\ \ \ \ \ \ \ \ \ \ \ "name": "param\_name", \\
\ \ \ \ \ \ \ \ \ \ \ "change\_type": "constraint|required|type", \\
\ \ \ \ \ \ \ \ \ \ \ "old\_value": "original\_value\_or\_type", \\
\ \ \ \ \ \ \ \ \ \ \ "new\_value": "new\_value\_or\_type" \\
\ \ \ \ \ \ \ \ \}, \\
\ \ \ \ \ \ \ \ ...(no more than three) \\
\ \ \ \ ], \\
\ \ \ \ \{Code\_Modification\_Examples\} \\
\}
\end{tcolorbox}

\begin{tcolorbox}[
breakable,
    enhanced,
    sharp corners, 
    boxrule=0.8pt, 
    colback=white,
    colframe=DeepBlackGray,
    colbacktitle=HeaderDarkGray,
    title=Prompt for Parameter Pruning Operator,
    fonttitle=\sffamily\bfseries,
    attach boxed title to top left={xshift=3mm, yshift=-3mm}, 
    boxed title style={
        arc=1.5mm, 
        boxrule=0pt,
        top=1mm, bottom=1mm, left=2mm, right=2mm
    },
    top=5mm, 
    left=4mm, right=4mm, bottom=4mm,
    fontupper=\ttfamily\small,
]
\#\# Objective \\
Simulate a tool evolution process by modifying the implementation code to remove redundant or low-value parameters. \\

\#\# Requirements \\
1. \textbf{Identify Candidates}: Select parameters for removal based on redundancy, or simplification goals. \\
2. \textbf{Remove Parameters}: Update the schema to remove optional or required parameters. \\
3. \textbf{Clean Logic}: Remove all code logic related to the deleted parameters (no dead code). \\
4. \textbf{Assess Breaking Changes}: Explicitly flag if removing a required parameter breaks existing clients. \\

\#\# Output Format \\
\{ \\
\ \ \ \ "evolution\_summary": "Brief overview of the pruning action and logic updates", \\
\ \ \ \ "parameter\_removals": [ \\
\ \ \ \ \ \ \ \ \{ \\
\ \ \ \ \ \ \ \ \ \ \ "name": "param\_name", \\
\ \ \ \ \ \ \ \ \ \ \ "type": "string|int|bool|...", \\
\ \ \ \ \ \ \ \ \ \ \ "required": true|false \\
\ \ \ \ \ \ \ \ \}, \\
\ \ \ \ \ \ \ \ ...(no more than three) \\
\ \ \ \ ], \\
\ \ \ \ \{Code\_Modification\_Examples\} \\
\}
\end{tcolorbox}

\begin{tcolorbox}[
breakable,
    enhanced,
    sharp corners, 
    boxrule=0.8pt, 
    colback=white,
    colframe=DeepBlackGray,
    colbacktitle=HeaderDarkGray,
    title=Prompt for Interface Refactoring Operator,
    fonttitle=\sffamily\bfseries,
    attach boxed title to top left={xshift=3mm, yshift=-3mm}, 
    boxed title style={
        arc=1.5mm, 
        boxrule=0pt,
        top=1mm, bottom=1mm, left=2mm, right=2mm
    },
    top=5mm, 
    left=4mm, right=4mm, bottom=4mm,
    fontupper=\ttfamily\small,
]
\#\# Objective \\
Simulate a tool evolution process where the \textbf{tool description} and \textbf{parameter structure} (add/remove) change simultaneously, ensuring documentation reflects the new interface shape. \\

\#\# Requirements \\
1. \textbf{Mandatory Tool Description Update}: The tool-level \texttt{description} must change to reflect the new capability or reduced scope. \\
2. \textbf{Parameter Structural Changes}: Include only \textbf{additions} or \textbf{removals} of parameters (no constraint/type modifications). \\
3. \textbf{Explicit Required Status}: Clearly indicate whether each changed parameter is \texttt{required} or \texttt{optional}. \\
4. \textbf{Description Sync}: Provide the description for added parameters (new) or removed parameters (old). \\

\#\# Output Format \\
\{ \\
\ \ \ \ "evolution\_summary": "Brief overview of the co-evolution", \\
\ \ \ \ "tool\_description\_change": \{ \\
\ \ \ \ \ \ \ \ "old\_description": "original tool description", \\
\ \ \ \ \ \ \ \ "new\_description": "updated tool description", \\
\ \ \ \ \ \ \ \ "reason": "why the tool description changed" \\
\ \ \ \ \}, \\
\ \ \ \ "parameter\_changes": [ \\
\ \ \ \ \ \ \ \ \{ \\
\ \ \ \ \ \ \ \ \ \ \ "name": "param\_name", \\
\ \ \ \ \ \ \ \ \ \ \ "action": "add|remove", \\
\ \ \ \ \ \ \ \ \ \ \ "required": true|false, \\
\ \ \ \ \ \ \ \ \ \ \ "description\_old": "original text (empty if add)", \\
\ \ \ \ \ \ \ \ \ \ \ "description\_new": "updated text (empty if remove)", \\
\ \ \ \ \ \ \ \ \ \ \ "reason": "why this parameter was added or removed" \\
\ \ \ \ \ \ \ \ \}...(no more than three) \\
\ \ \ \ ], \\
\ \ \ \ \{Code\_Modification\_Examples\} \\
\} \\

\textbf{Design Note}: \texttt{parameter\_changes} focuses strictly on structural additions or removals. The \texttt{required} field is mandatory to clarify contract implications. \texttt{tool\_description\_change} is mandatory to ensure high-level documentation stays in sync with the parameter list.
\end{tcolorbox}

\begin{tcolorbox}[
breakable,
    enhanced,
    sharp corners, 
    boxrule=0.8pt, 
    colback=white,
    colframe=DeepBlackGray,
    colbacktitle=HeaderDarkGray,
    title=Prompt for Tool Description Update Operator,
    fonttitle=\sffamily\bfseries,
    attach boxed title to top left={xshift=3mm, yshift=-3mm}, 
    boxed title style={
        arc=1.5mm, 
        boxrule=0pt,
        top=1mm, bottom=1mm, left=2mm, right=2mm
    },
    top=5mm, 
    left=4mm, right=4mm, bottom=4mm,
    fontupper=\ttfamily\small,
]
You are an expert Technical Writer specializing in MCP (Model Context Protocol) documentation. Your task is to simulate the iterative evolution of an MCP Server by rewriting the tool description to accurately reflect its functionality. \\

\#\#\# Context \\
\textbf{Server Name:} \{SERVER\_NAME\} \\
\textbf{Server Description:} \{SERVER\_DESC\} \\
\textbf{Target Tool Status:} \\
\{TARGET\_TOOL\_STATUS\} \\
(Note: This input contains the specific target tool's details, including current tool's \texttt{name}, \texttt{description}, \texttt{inputSchema}, \texttt{outputSchema}) \\

\#\#\# Critical Constraints \\

1. \textbf{Schema Immunity}: \\
You MUST NOT alter the tool's \texttt{name}, \texttt{inputSchema}, \texttt{outputSchema}, or any internal logic. Your modifications are \textbf{strictly limited to the \texttt{description} text field} of the target tool. The structural integrity of the tool definition must remain intact. \\

2. \textbf{Precision \& Consistency}: \\
Rewrite the \textbf{Target Tool Description} based \textit{only} on the provided \texttt{inputSchema} and \texttt{outputSchema}. \\
- \textbf{Strict Prohibition}: Do not claim, imply, or promise any capabilities, parameters, or return values that are not explicitly defined in the provided schemas. Avoid hallucinating features or ignoring constraints (e.g., if a parameter is optional in the schema, do not describe it as mandatory). \\

3. \textbf{Output Format Requirement}: \\
- You must output \textbf{ONLY} a single, valid JSON object. \\
- The JSON object must contain exactly two keys: \texttt{original\_description}, and \texttt{modified\_description}. \\
\{ \\
\ \ "tool": \{ \\
\ \ \ \ "original\_description": "The exact value of the current tool description", \\
\ \ \ \ "modified\_description": "the newly generated tool description" \\
\ \ \} \\
\} \\
- \textbf{NO Markdown}: Do not wrap the output in markdown code blocks. Do not include any explanations, introductory text, or trailing comments. The output must be a raw JSON string directly parsable by a machine. \\
- Ensure the language of the modifications remains consistent with the original text whenever possible.
\end{tcolorbox}

\begin{tcolorbox}[
breakable,
    enhanced,
    sharp corners, 
    boxrule=0.8pt, 
    colback=white,
    colframe=DeepBlackGray,
    colbacktitle=HeaderDarkGray,
    title=Prompt for Parameters Description Update Operator,
    fonttitle=\sffamily\bfseries,
    attach boxed title to top left={xshift=3mm, yshift=-3mm}, 
    boxed title style={
        arc=1.5mm, 
        boxrule=0pt,
        top=1mm, bottom=1mm, left=2mm, right=2mm
    },
    top=5mm, 
    left=4mm, right=4mm, bottom=4mm,
    fontupper=\ttfamily\small,
]
You are an expert Technical Writer specializing in MCP (Model Context Protocol) documentation. Your task is to simulate the iterative evolution of an MCP Server by rewriting the descriptions of specific parameters within a tool's \texttt{properties}. \\

\#\#\# Context \\
\textbf{Server Name:} \{SERVER\_NAME\} \\
\textbf{Server Description:} \{SERVER\_DESC\} \\
\textbf{Target Tool Status:} \\
\{TARGET\_TOOL\_STATUS\} \\
(Note: This input contains the specific target tool's details, including current tool's \texttt{name}, \texttt{description}, \texttt{inputSchema}, \texttt{outputSchema}) \\

\#\#\# Critical Constraints \\

1. \textbf{Deep Schema Immunity}: \\
- You MUST NOT alter the \texttt{name}, \texttt{type}, \texttt{format}, \texttt{enum}, \texttt{required} status in \texttt{inputSchema}, or any structural definition of the parameters. \\
- You MUST NOT add new parameters or remove existing ones. \\
- Your modifications are \textbf{strictly limited to the \texttt{description} string} inside the \texttt{inputSchema} for the specified parameters. \\
- The rest of the tool configuration (tool description, outputSchema, other unchanged parameters) must remain exactly as provided in the source context. \\

2. \textbf{Precision \& Consistency}: \\
- Rewrite the target parameter descriptions to precisely match their data types and constraints defined in the schema. \\
- \textbf{Strict Prohibition}: Do not change the semantic meaning of the parameter in a way that contradicts its type or validation rules. Do not hallucinate constraints that do not exist. \\

3. \textbf{Output Format Requirement}: \\
- Output \textbf{ONLY} a raw JSON: \\
\{ \\
\ \ \ \ "parameters": [ \\
\ \ \ \ \ \ \ \ \{ \\
\ \ \ \ \ \ \ \ \ \ \ "parameter\_name": "the\_name\_of\_the\_parameter", \\
\ \ \ \ \ \ \ \ \ \ \ "original\_description": "the exact input description before modification", \\
\ \ \ \ \ \ \ \ \ \ \ "modified\_description": "the newly generated description" \\
\ \ \ \ \ \ \ \ \} \\
\ \ \ \ \ \ \ \ ... \\
\ \ \ \ ] \\
\} \\
- \textbf{NO Markdown}: Do not wrap the output in markdown code blocks. Do not include any explanations, introductory text, or trailing comments. The output must be a raw JSON list string directly parsable by a machine. \\
- Ensure the language of the modifications remains consistent with the original text whenever possible.
\end{tcolorbox}

\begin{tcolorbox}[
breakable,
    enhanced,
    sharp corners, 
    boxrule=0.8pt, 
    colback=white,
    colframe=DeepBlackGray,
    colbacktitle=HeaderDarkGray,
    title=Prompt for Joint Description Update Operator,
    fonttitle=\sffamily\bfseries,
    attach boxed title to top left={xshift=3mm, yshift=-3mm}, 
    boxed title style={
        arc=1.5mm, 
        boxrule=0pt,
        top=1mm, bottom=1mm, left=2mm, right=2mm
    },
    top=5mm, 
    left=4mm, right=4mm, bottom=4mm,
    fontupper=\ttfamily\small,
]
You are an expert Technical Writer specializing in MCP (Model Context Protocol) documentation. Your task is to simulate the iterative evolution of an MCP Server by rewriting the descriptions of specific parameters within a tool's \texttt{properties} and the tool's description, ensuring strictly semantic adjustment without logic changes. \\

\#\#\# Context \\
\textbf{Server Name:} \{SERVER\_NAME\} \\
\textbf{Server Description:} \{SERVER\_DESC\} \\
\textbf{Target Tool Status:} \\
\{TARGET\_TOOL\_STATUS\} \\
(Note: This input contains the specific target tool's details, including current tool's \texttt{name}, \texttt{description}, \texttt{inputSchema}, \texttt{outputSchema}) \\

\#\#\# Core Rule: SEMANTIC ONLY, NO LOGIC CHANGES \\
- \textbf{DO}: Improve clarity, grammar, tone, conciseness. Clarify \textit{existing} constraints found in the schema. \\
- \textbf{DON'T}: Add features, change types/format/requirements, or imply capabilities not in the schema. \\
- \textbf{Critical}: If schema says "optional", description cannot say "required". \\

\#\#\# Output Requirement \\
- Output \textbf{ONLY} a raw JSON: \\
\{ \\
\ \ \ \ "tool": \{ \\
\ \ \ \ \ \ \ \ "original\_description": "The exact value of the current tool description", \\
\ \ \ \ \ \ \ \ "modified\_description": "the newly generated tool description" \\
\ \ \ \ \}, \\
\ \ \ \ "parameters": [ \\
\ \ \ \ \ \ \ \ \{ \\
\ \ \ \ \ \ \ \ \ \ \ "parameter\_name": "the\_name\_of\_the\_parameter", \\
\ \ \ \ \ \ \ \ \ \ \ "original\_description": "the exact input description before modification", \\
\ \ \ \ \ \ \ \ \ \ \ "modified\_description": "the newly generated description" \\
\ \ \ \ \ \ \ \ \} \\
\ \ \ \ \ \ \ \ ... \\
\ \ \ \ ] \\
\} \\
- \textbf{NO Markdown}: Do not wrap the output in markdown code blocks. Do not include any explanations, introductory text, or trailing comments. The output must be a raw JSON list string directly parsable by a machine. \\
- Ensure the language of the modifications remains consistent with the original text whenever possible.
\end{tcolorbox}

\begin{tcolorbox}[
    breakable,
    enhanced,
    sharp corners, 
    boxrule=0.8pt, 
    colback=white,
    colframe=DeepBlackGray,
    colbacktitle=HeaderDarkGray,
    title=Prompt for File Modification Instruction,
    fonttitle=\sffamily\bfseries,
    attach boxed title to top left={xshift=3mm, yshift=-3mm}, 
    boxed title style={
        arc=1.5mm, 
        boxrule=0pt,
        top=1mm, bottom=1mm, left=2mm, right=2mm
    },
    top=5mm, 
    left=4mm, right=4mm, bottom=4mm,
    fontupper=\ttfamily\small,
]
\# Role \\
You are an MCP Server Code Modification Expert. Based on the provided modification requirements, output the content of the modified TypeScript code. \\

\# Input \\
- \textbf{Server Name}: \{SERVER\_NAME\} \\
- \textbf{Server Description}: \{SERVER\_DESC\} \\
- \textbf{Target Tool}: \{TARGET\_TOOL\_NAME\} \\
- \textbf{TypeScript Code}: \{ORIGINAL\_CODE\} \\

\# Modification Requirement \\
\{Operator\_FUNCTION\} \\

\# Constraints \\
1. \textbf{Minimal Changes}: Only modify the logic related to the Target Tool implementation, preserve all other code exactly as is. \\
2. \textbf{Maximal Enhancements}: Implement extensive functional and contextual modifications, while strictly preserving the core original functionality of the MCP server. \\
3. \textbf{Valid Syntax}: Ensure the output TypeScript code is compilable and runnable. \\
4. \textbf{Valid Logic}: Changes are based on the original logic, meaning evolution is achieved by leveraging the code logic of existing tools. Do not fabricate information sources, such as accessing self-invented URLs in the code or using data with unknown formats or structures. \\
5. \textbf{Crucial}: Make sure not to output the modified complete code. Wrap the entire JSON content inside Markdown code blocks with the \texttt{json} language identifier. \\
- Format: \\
\{ \\
\ \ "key": "value" \\
\} \\
\end{tcolorbox}

\subsection{Prompts for Benchmark Construction}
\label{app:prompt:construction}
Presented below are the prompts utilized for three key tasks: extracting MCP server installation configurations, server function classification, multi-server task generation.

\begin{tcolorbox}[
    breakable,
    enhanced,
    sharp corners, 
    boxrule=0.8pt, 
    colback=white,
    colframe=DeepBlackGray,
    colbacktitle=HeaderDarkGray,
    title=Prompt for MCP Installation Config Extraction,
    fonttitle=\sffamily\bfseries,
    attach boxed title to top left={xshift=3mm, yshift=-3mm}, 
    boxed title style={
        arc=1.5mm, 
        boxrule=0pt,
        top=1mm, bottom=1mm, left=2mm, right=2mm
    },
    top=5mm, 
    left=4mm, right=4mm, bottom=4mm,
    fontupper=\ttfamily\small,
]
You are an expert MCP (Model Context Protocol) configuration extractor. Analyze the provided README content and generate a \textbf{strictly valid JSON output} with two top-level properties: \\

1. \texttt{"config"} -- VS Code-compatible MCP server configuration object \\
2. \texttt{"metadata"} -- API key requirement analysis \\

\textbf{Extraction Rules:} \\

\checkmark \textbf{For \texttt{config.mcpServers}:} \\
- Identify the primary MCP server name from README (e.g., package name, title, or explicit server name) \\
- Extract launch command pattern (e.g., \texttt{npx <package>}, \texttt{python -m server}, \texttt{docker run...}) \\
- Split command into: \\
\ \ - \texttt{"command"}: Executable only (e.g., \texttt{"npx"}, \texttt{"python"}, \texttt{"docker"}) \\
\ \ - \texttt{"args"}: Array of arguments (e.g., \texttt{["bazi-mcp"]}, \texttt{["-m", "my\_server"]}) \\
- If multiple servers exist, include all under \texttt{mcpServers} with unique keys \\
- If command pattern is ambiguous, use the most prominently documented launch method \\
- Default to empty object \texttt{\{\}} for \texttt{mcpServers} if no launch instructions found \\

\checkmark \textbf{For \texttt{metadata.requires\_api\_key}:} \\
- \texttt{true} ONLY if README explicitly states external API credentials are \textbf{required for basic operation} (e.g., "You must set OPENAI\_API\_KEY to start the server") \\
- \texttt{false} if: \\
\ \ - Server works locally without external services \\
\ \ - API keys are only for optional features \\
\ \ - No credential requirements mentioned \\
- Be conservative -- prefer \texttt{false} when uncertain \\

\checkmark \textbf{For \texttt{metadata.api\_key\_examples}:} \\
- List ONLY environment variable names explicitly mentioned as API keys (e.g., \texttt{["GROQ\_API\_KEY", "ANTHROPIC\_API\_KEY"]}) \\
- Empty array \texttt{[]} if none detected \\

\textbf{Output Schema (STRICTLY VALID JSON):} \\
\{ \\
\ \ \ \ "config": \{ \\
\ \ \ \ \ \ \ \ "mcpServers": \{ \\
\ \ \ \ \ \ \ \ \ \ \ "ServerName": \{ \\
\ \ \ \ \ \ \ \ \ \ \ \ \ \ "command": "string (npx command)", \\
\ \ \ \ \ \ \ \ \ \ \ \ \ \ "args": ["string"] \\
\ \ \ \ \ \ \ \ \ \ \ \} \\
\ \ \ \ \ \ \ \ \} \\
\ \ \ \ \}, \\
\ \ \ \ "metadata": \{ \\
\ \ \ \ \ \ \ \ "requires\_api\_key": boolean, \\
\ \ \ \ \ \ \ \ "api\_key\_examples": ["string"] \\
\ \ \ \ \} \\
\} \\

\textbf{Critical Requirements:} \\
- NEVER omit top-level keys (\texttt{config}, \texttt{metadata}) \\
- \texttt{mcpServers} must be an object (not array) with server names as keys \\
- Output ONLY raw JSON -- no explanations, markdown, or prefixes \\
- Escape special characters properly for valid JSON \\
- If no server detected: \texttt{"mcpServers": \{\}} \\
- If no API keys mentioned: \texttt{"api\_key\_examples": []} \\

\textbf{Input README:} \\
\{readme\_content\} \\
\end{tcolorbox}

\begin{tcolorbox}[
    breakable,
    enhanced,
    sharp corners, 
    boxrule=0.8pt, 
    colback=white,
    colframe=DeepBlackGray,
    colbacktitle=HeaderDarkGray,
    title=Prompt for MCP Server Categorization,
    fonttitle=\sffamily\bfseries,
    attach boxed title to top left={xshift=3mm, yshift=-3mm}, 
    boxed title style={
        arc=1.5mm, 
        boxrule=0pt,
        top=1mm, bottom=1mm, left=2mm, right=2mm
    },
    top=5mm, 
    left=4mm, right=4mm, bottom=4mm,
    fontupper=\ttfamily\small,
]
You are a system architecture expert. Based on the following MCP server descriptions, define a set of high-level functional categories (e.g., Finance, File System, Browser, Authentication, etc.), assign each server to one or more categories, and provide a short description for each category. \\

\# Requirements: \\
1. Define intuitive, meaningful category names that reflect broad functional domains. \\
2. The assigned categories should be specific to a particular field and not vague or multiple fields. \\
3. Use \textbf{no more than 10 distinct categories} in total. \\
4. Aim to distribute the servers \textbf{as evenly as possible} across the categories. \\
5. Each server must be assigned to one category. \\
6. Output a valid JSON object with three keys: \\
- \texttt{"categories"}: a list of objects, each containing \texttt{"name"} and \texttt{"description"} fields; \\
- \texttt{"assignments"}: a mapping from each server name to a list of category names it belongs to; \\
- The \texttt{"description"} for each category should be a concise sentence (10–20 words) explaining what kinds of services belong in it. \\

Example output structure: \\
\{ \\
\ \ \ \ "categories": [ \\
\ \ \ \ \ \ \ \ \{"name": "Finance", "description": "Services related to payments, billing, and financial transactions."\}, \\
\ \ \ \ \ \ \ \ \{"name": "FileSystem", "description": "Operating system files, documents, and other related content."\} \\
\ \ \ \ ], \\
\ \ \ \ "assignments": \{ \\
\ \ \ \ \ \ \ \ "payment-service": ["Finance"], \\
\ \ \ \ \ \ \ \ "file-converter": ["FileSystem"] \\
\ \ \ \ \} \\
\} \\

MCP servers to classify: \\
\{services\_list\}
\end{tcolorbox}

\begin{tcolorbox}[
    breakable,
    enhanced,
    sharp corners, 
    boxrule=0.8pt, 
    colback=white,
    colframe=DeepBlackGray,
    colbacktitle=HeaderDarkGray,
    title=Prompt for Multi-Server Task Generation,
    fonttitle=\sffamily\bfseries,
    attach boxed title to top left={xshift=3mm, yshift=-3mm}, 
    boxed title style={
        arc=1.5mm, 
        boxrule=0pt,
        top=1mm, bottom=1mm, left=2mm, right=2mm
    },
    top=5mm, 
    left=4mm, right=4mm, bottom=4mm,
    fontupper=\ttfamily\small,
]
\#\# Task \\
Generate a realistic user question that requires tools from \textbf{multiple MCP servers} to resolve. \\

\#\# Guidelines \\
- The question must be concise and should reflect a \textbf{practical, real-world workflow} involving \textbf{different servers (>{}=2)}. \\
- Include sufficient contextual information to \textbf{unambiguously determine all needed tool calls and their parameters}, but do not use real user names. \\
- Phrase it naturally—as if asked by a real user—\textbf{without mentioning any tool names, server names, or technical internals}. \\
- The solution should require \textbf{multiple tool invocations across servers}. \\
- \textbf{Do not use all servers/tools}—instead, select a \textbf{coherent subset} that covers distinct capabilities within one realistic scenario. \\
- Rely \textbf{only on the provided servers and tools}—no external knowledge or assumptions. \\
- The question may be written in English or in the same language as the tool descriptions. \\
- Regarding real-time issues, the time is set around \{set\_time\}. \\
- Cannot use files that do not exist locally unless you have previously saved them. \\

\#\# Input \\
Available MCP Servers: \\
\{SERVER\_DESCRIPTIONS\} \\

Relevant files (if applicable): \\
\{FILE\_EXISTS\} \\

Database Configuration (if applicable): \\
- Host: 127.0.0.1 \\
- Port: 3305 \\
- User: root \\
- Password: 123456 \\
- Database Name: mydb \\

\#\# Output Format \\
Return exactly one response in the following XML structure: \\

\textless response\textgreater \\
\ \ \ \ \textless workflow\_analysis\textgreater \\
\ \ \ \ \ \ \ \ \% Briefly explain how the selected servers interact in this scenario \\
\ \ \ \ \textless /workflow\_analysis\textgreater \\
\ \ \ \ \textless target\_servers\textgreater \\
\ \ \ \ \ \ \ \ \textless server name="server\_name\_1"\textgreater \\
\ \ \ \ \ \ \ \ \ \ \ \ \textless tool\textgreater tool\_name\_a\textless /tool\textgreater \\
\ \ \ \ \ \ \ \ \ \ \ \ \textless tool\textgreater tool\_name\_b\textless /tool\textgreater \\
\ \ \ \ \ \ \ \ \textless /server\textgreater \\
\ \ \ \ \ \ \ \ \textless server name="server\_name\_2"\textgreater \\
\ \ \ \ \ \ \ \ \ \ \ \ \textless tool\textgreater tool\_name\_c\textless /tool\textgreater \\
\ \ \ \ \ \ \ \ \textless /server\textgreater \\
\ \ \ \ \ \ \ \ \% Add more servers/tools as needed \\
\ \ \ \ \textless /target\_servers\textgreater \\
\ \ \ \ \textless question\textgreater \\
\ \ \ \ \ \ \ \ \% Natural-language user question \\
\ \ \ \ \textless /question\textgreater \\
\textless /response\textgreater
\end{tcolorbox}

\subsection{Prompts for Benchmark Evaluation}
\label{app:prompt:eval}
We provide prompts for LLM-based task execution, rubric-based trajectory evalution and trajectory error type identification.

\begin{tcolorbox}[
    breakable,
    enhanced,
    sharp corners, 
    boxrule=0.8pt, 
    colback=white,
    colframe=DeepBlackGray,
    colbacktitle=HeaderDarkGray,
    title=Prompt for Task Execution,
    fonttitle=\sffamily\bfseries,
    attach boxed title to top left={xshift=3mm, yshift=-3mm}, 
    boxed title style={
        arc=1.5mm, 
        boxrule=0pt,
        top=1mm, bottom=1mm, left=2mm, right=2mm
    },
    top=5mm, 
    left=4mm, right=4mm, bottom=4mm,
    fontupper=\ttfamily\small,
]
You are an AI agent connected to the MCP (Model Context Protocol) server. Based on the user's request, select and invoke tools to fulfill the task accurately and efficiently. \\
Note that do not to ask any questions to the user, just call the tool to complete the task. \\

Considering real-time performance, the time is set to be around \{SET\_TIME\}. \\

These files are stored in the local file system and serve as relevant resources for tool utilization. Add the prefix path \texttt{./anotation\_path} to the following file systems as an absolute path during use: \\
\texttt{\{FILE\_EXISTS\}} \\

Database Configuration (if applicable): \\
- Host: 127.0.0.1 \\
- Port: 3305 \\
- User: root \\
- Password: 123456 \\
- Database Name: mydb
\end{tcolorbox}

\begin{tcolorbox}[
    breakable,
    enhanced,
    sharp corners, 
    boxrule=0.8pt, 
    colback=white,
    colframe=DeepBlackGray,
    colbacktitle=HeaderDarkGray,
    title=Prompt for Agent Trajectory Evaluation,
    fonttitle=\sffamily\bfseries,
    attach boxed title to top left={xshift=3mm, yshift=-3mm}, 
    boxed title style={
        arc=1.5mm, 
        boxrule=0pt,
        top=1mm, bottom=1mm, left=2mm, right=2mm
    },
    top=5mm, 
    left=4mm, right=4mm, bottom=4mm,
    fontupper=\ttfamily\small,
]
\# System Role \\
You are an impartial evaluator judging the quality of an AI agent's multi-server tool-based task execution. You focus on three independent dimensions: Task Completion, Grounding, and Planning Efficiency. \\

\# User Instructions \\
You must assign scores based ONLY on evidence from the task, solution, and tool usage trajectory. \\
- Objective: Ignore language fluency, formatting, or politeness. \\
- Justified: Every score must be backed by specific counts/percentages derived from the trajectory. \\
- Independent: Evaluate each dimension separately. A flaw in one dimension (e.g., redundancy) should NOT lower the score of another (e.g., task completion) unless it directly caused failure. \\

\# Input Data \\
TASK PRESENTED TO AGENT: \\
\{task\} \\

AVAILABLE TOOLS: \\
\{available\_tools\} \\

AGENT TRAJECTORY FOR TASK COMPLETION: \\
\{trajectory\} \\

FINAL ANSWER \\
\{answer\} \\

\# SCORING RUBRICS (1–10 PER SUBDIMENSION) \\
Scores are derived from the DEFECT RATE calculated in the principles below. \\

\#\# Task Fulfillment \\
Measures: Did the agent achieve the user's explicit goals? \\
- 1–3: Perfectly completes 10–30\% of requirements. \\
- 4–6: Perfectly completes 40–60\% of requirements. \\
- 7–8: Perfectly completes 70–80\% of requirements. \\
- 9–10: Perfectly completes 90–100\% of requirements. \\

\#\# Information Grounding \\
Measures: Are the agent's final assertions supported by tool outputs? \\
- 1–3: 10–30\% of claims are perfectly grounded in tool outputs. \\
- 4–6: 40–60\% of claims are perfectly grounded in tool outputs. \\
- 7–8: 70–80\% of claims are perfectly grounded in tool outputs. \\
- 9–10: 90–100\% of claims are perfectly grounded in tool outputs. \\

\#\# Planning Efficiency \\
Measures: Did the agent avoid redundant or unnecessary tool calls? \\
- 9–10: \textless 10\% of calls were redundant/unnecessary. \\
- 7–8: 10–30\% of calls were redundant/unnecessary. \\
- 4–6: 30–60\% of calls were redundant/unnecessary. \\
- 1–3: \textgreater 60\% of calls were redundant/unnecessary. \\

\# CALCULATION METHODOLOGY \\

\#\# Step 1: Define "Opportunities" (Denominator) for Each Dimension \\
\textbf{Task Fulfillment}: Count the distinct, explicit sub-goals/requirements in the \textbf{TASK PRESENTED TO AGENT}. \\
\textbf{Grounding}: Count the atomic factual claims or data points in the agent's FINAL response. \\
\textbf{Planning Efficiency}: Count the TOTAL number of tool calls made in the \textbf{AGENT TRAJECTORY FOR TASK COMPLETION}. \\

\#\# Step 2: Identify "Issues" (Numerator) for Each Dimension \\
\textbf{Task Fulfillment Issues}: Sub-goals that were FAILED, IGNORED, or INCORRECTLY executed. \\
- Note: If a sub-goal was met but via a redundant path, it is STILL COUNTED AS MET for this dimension. Do not penalize efficiency here. \\
\textbf{Grounding Issues}: Claims in the final answer that are NOT found in or CONTRADICT the tool outputs. \\
\textbf{Planning Efficiency Issues}: Tool calls that were REDUNDANT (repeating previous successful calls), UNNECESSARY (not needed for any sub-goal), or RETRIED without parameter changes after failure. \\
- Note: Necessary calls that were just "slow" or "sub-optimal" but not redundant are NOT issues. \\

\#\# Step 3: Calculate Defect Rate \& Map to Score \\
- Defect Rate = (Issues / Opportunities) $\times$ 100\% \\
- Mapping: \\
- 0–10\% defects $\rightarrow$ Score 9–10 \\
- 10–30\% defects $\rightarrow$ Score 7–8 \\
- 30–60\% defects $\rightarrow$ Score 4–6 \\
- 60–100\% defects $\rightarrow$ Score 1–3 \\

\# How to Score: \\
1. When evaluating percentages, be EXTREMELY STRICT about what counts as "perfectly executed". \\
2. "Perfectly" means ALL of the following must be true: \\
- Complete and accurate parameters (not just valid, but IDEAL) \\
- Zero redundancy (no repeated or unnecessary calls) \\
- Proper error handling (graceful recovery from ANY failure) \\
- Efficient execution (minimal rounds) \\
3. If ANY of the above is missing, that portion is NOT perfectly executed (counts as 0\%). \\
4. Example: Task completed correctly but with 1 redundant call = that portion is 0\% perfect. \\

\# KEY PRINCIPLES: \\
1. ALWAYS calculate as percentage, NOT absolute numbers. \\
2. 10 errors in 100 calls (10\%) = same score as 1 error in 10 calls (10\%). \\
3. NORMALIZE by complexity - don’t punish complex tasks: \\
- Simple task: 1 error/5 steps (20\% defect) = Score 7 \\
- Complex task: 4 errors/20 steps (20\% defect) = Score 7 \\
4. CRITICAL: Apply the STRICTEST interpretation of “perfectly executed”. If there’s ANY doubt, score lower. \\

\# FINAL REMINDER BEFORE SCORING: \\
- Count ONLY truly perfect executions toward the percentage \\
- Be your most critical self - find flaws first, then acknowledge successes \\
- If you're considering a score above 8, re-examine for ANY imperfection \\
- Server count is IRRELEVANT - using more servers is NOT better \\

\# OUTPUT FORMAT \\
Return your evaluation scoring and reasoning in this exact JSON format. Return ONLY the JSON object. \\
\{ \\
\ \ \ \ "task\_fulfillment\_reasoning": "List total requirements vs. met requirements. Calculate defect rate.", \\
\ \ \ \ "grounding\_reasoning": "List total claims vs. unsupported claims. Calculate defect rate.", \\
\ \ \ \ "planning\_efficiency\_reasoning": "List total tool calls vs. redundant/unnecessary calls. Calculate defect rate.", \\
\ \ \ \ "task\_fulfillment": \textless integer\_score\textgreater, \\
\ \ \ \ "grounding": \textless integer\_score\textgreater, \\
\ \ \ \ "planning\_and\_efficiency": \textless integer\_score\textgreater \\
\} \\
\end{tcolorbox}

\begin{tcolorbox}[
    breakable,
    enhanced,
    sharp corners, 
    boxrule=0.8pt, 
    colback=white,
    colframe=DeepBlackGray,
    colbacktitle=HeaderDarkGray,
    title=Prompt for Test Case Generation,
    fonttitle=\sffamily\bfseries,
    attach boxed title to top left={xshift=3mm, yshift=-3mm}, 
    boxed title style={
        arc=1.5mm, 
        boxrule=0pt,
        top=1mm, bottom=1mm, left=2mm, right=2mm
    },
    top=5mm, 
    left=4mm, right=4mm, bottom=4mm,
    fontupper=\ttfamily\small,
]
\# Server \\
name: \{server\_name\} \\
server\_description: \{server\_description\} \\

\# Registered tool (schema summary) \\
\{tool\_schema\} \\

\# Task \\
Generate multiple test cases to maximize parameter coverage and validate all constraints for the tool provided above. \\

\# Rules \\
- Standard Execution: Provide cases that include all parameters (required + optional) and cases with only the minimum required fields. \\
- Boundary Coverage: Test numeric ranges (min/max), string length limits, and specific enum values. \\
- Validation Testing: Include cases with missing required fields, incorrect data types, or strings violating regex patterns. \\
- Realism: Values in \texttt{input\_arguments} must be semantically appropriate for the tool's specific functionality. \\

Output ONLY valid JSON (no markdown fences): \\
\{ \\
\ \ \ \ "tool\_name": string, \\
\ \ \ \ "test\_cases": [ \\
\ \ \ \ \ \ \ \ \{ \\
\ \ \ \ \ \ \ \ \ \ \ "test\_name": string, \\
\ \ \ \ \ \ \ \ \ \ \ "target\_parameters": [string], \\
\ \ \ \ \ \ \ \ \ \ \ "input\_arguments": \{ \\
\ \ \ \ \ \ \ \ \ \ \ \ \ \ "parameter\_name\_1": "value\_1", \\
\ \ \ \ \ \ \ \ \ \ \ \ \ \ "parameter\_name\_2": "value\_2", \\
\ \ \ \ \ \ \ \ \ \ \ \ \ \ ... \\
\ \ \ \ \ \ \ \ \ \ \ \}, \\
\ \ \ \ \ \ \ \ \ \ \ "expected\_status": "Success" | "Validation Error", \\
\ \ \ \ \ \ \ \ \ \ \ "objective": "description of the specific parameter or constraint being tested" \\
\ \ \ \ \ \ \ \ \} \\
\ \ \ \ ] \\
\}
\end{tcolorbox}

\begin{tcolorbox}[
    breakable,
    enhanced,
    sharp corners, 
    boxrule=0.8pt, 
    colback=white,
    colframe=DeepBlackGray,
    colbacktitle=HeaderDarkGray,
    title=Prompt for Trajectory Error Type Identification,
    fonttitle=\sffamily\bfseries,
    attach boxed title to top left={xshift=3mm, yshift=-3mm}, 
    boxed title style={
        arc=1.5mm, 
        boxrule=0pt,
        top=1mm, bottom=1mm, left=2mm, right=2mm
    },
    top=5mm, 
    left=4mm, right=4mm, bottom=4mm,
    fontupper=\ttfamily\small,
]
\# Role \\
You are an expert analyst classifying \textbf{LLM MCP agent trajectories} into exactly one \textbf{primary error category} according to the six-dimensional orthogonal taxonomy below. \\

\# Taxonomy (Choose the single best category) \\

1. \textbf{specification\_syntax\_error} \\
Technical violations of the tool-calling protocol. This includes: \\
- Invalid parameter names or other syntax errors. \\
- Missing required parameters or incorrect parameter names. \\
- Calling tools that exist but providing values that violate the API schema/contract. \\

2. \textbf{intent\_tool\_misalignment} \\
Selecting an inappropriate tool for the task from the available set. This includes: \\
- Failing to choose the correct tool to resolve the user's intent. \\
- Choosing a tool that does not align with the current sub-task. \\

3. \textbf{semantic\_parameter\_error} \\
The tool and format are correct, but the parameter values are semantically wrong. This includes: \\
- Semantic shifts (e.g., swapping identities, wrong dates). \\
- Value hallucinations (filling info not grounded in context). \\
- Missing critical information extraction or incorrect value types. \\

4. \textbf{planning\_logic\_error} \\
Errors in the multi-step workflow orchestration. This includes: \\
- Missing preconditions (e.g., acting before searching). \\
- Infinite loops or getting stuck in a logic cycle. \\
- Incomplete steps or prematurely ending the task. \\

5. \textbf{operational\_redundancy} \\
The task might succeed, but the path is inefficient or wasteful. This includes: \\
- Repeatedly calling the same tool with identical inputs. \\
- Performing unnecessary actions that do not contribute to the final result. \\

6. \textbf{reasoning\_observation\_error} \\
Failure to correctly interpret or react to tool outputs (Observations). This includes: \\
- Misinterpreting structured data or status codes returned by the API. \\
- Ignoring error messages and proceeding with false assumptions. \\

\# Rules \\
1. Pick exactly \textbf{one} \texttt{primary\_category} from the list above. \\
2. If the trajectory is perfectly correct and efficient, set \texttt{primary\_category} to \texttt{"none"}. \\
3. If multiple issues appear, choose the \textbf{earliest root cause} that led the agent astray. \\
4. Base your judgment on the provided Task, Tools, Final Answer, and Trajectory. \\
5. Select \texttt{operational\_redundancy} only if you are sure that no other error types are present. \\

\# Input \\

\#\# Task (User Question) \\
\{QUESTION\} \\

\#\# Available Tools \\
\{SERVERS\_TOOLS\} \\

\#\# Final Answer \\
\{FINAL\_ANSWER\} \\

\#\# Trajectory \\
\{TRAJECTORY\} \\

\#\# Optional Evaluation Hints \\
\{EVAL\_HINTS\} \\

\# Output (JSON only) \\
Return ONLY a valid JSON object. \\

\{ \\
\ \ \ \ "primary\_category": "\textless spec\_syntax\_err| intent\_misalign| sem\_param\_err| plan\_logic\_err| op\_redundancy| reason\_obs\_err| none\textgreater", \\
\ \ \ \ "confidence": \textless number from 0 to 1\textgreater, \\
\ \ \ \ "evidence": "\textless 1-3 short sentences citing specific steps where the error occurred\textgreater", \\
\ \ \ \ "analysis": "\textless detailed rationale in English or Chinese explaining why this specific category was chosen\textgreater" \\
\}
\end{tcolorbox}


\end{document}